\documentclass{article} 
\usepackage{iclr2027_conference,times}

\usepackage{amsmath,amsfonts,bm}

\def\eqref#1{equation~\ref{#1}}

\def\1{\bm{1}}

\DeclareMathAlphabet{\mathsfit}{\encodingdefault}{\sfdefault}{m}{sl}
\SetMathAlphabet{\mathsfit}{bold}{\encodingdefault}{\sfdefault}{bx}{n}

\usepackage{hyperref}
\usepackage{url}

\title{\textnormal{\textbf{Dyad}}: Extending Large Language Models with Native Typed Decision-Making}

\author{
\textbf{
Yundaichuan Zhan$^{1}$ \quad
Weishi Wang$^{2}$ \quad
Wenbiao Liu$^{3}$ \quad
Daniel Dahlmeier$^{2}$
}\\
\textbf{
Chengwei Qin$^{4}$ \quad
Juncheng Li$^{1}$\thanks{Corresponding author.} \quad
Fredrik D. Johansson$^{5}$ \quad
Zhongqi Yue$^{6}$
}\\[3pt]
$^{1}$Zhejiang University \quad
$^{2}$SAP \quad
$^{3}$Central South University\\
$^{4}$The Hong Kong University of Science and Technology (Guangzhou)\\
$^{5}$Chalmers University of Technology \quad
$^{6}$Microsoft Research
}

\usepackage{amsmath}
\usepackage{enumitem}
\usepackage{amsthm}
\usepackage{wrapfig}
\usepackage{graphicx}
\usepackage{algorithm,algpseudocode}
\usepackage{subcaption}
\usepackage{dsfont}
\usepackage{mathtools}
\usepackage{tabularx}
\usepackage{amssymb}
\usepackage[dvipsnames]{xcolor}
\usepackage{etoc} 

\newcommand{\ie}{\textit{i.e.}}
\newcommand{\eg}{\textit{e.g.}}

\usepackage[T1]{fontenc}
\usepackage{booktabs}
\usepackage{multirow}
\usepackage{tcolorbox}
\usepackage{float}
\usepackage{soul}   
\usepackage{pifont} 

\definecolor{PromptFrame}{HTML}{34495E}
\definecolor{PromptFill}{HTML}{F7F8FA}
\definecolor{PromptVariable}{HTML}{1455A0}
\definecolor{PromptSyntax}{HTML}{9C2348}

\newtcolorbox{PromptTemplate}[1]{
  title={#1},
  colback=PromptFill,
  colframe=PromptFrame,
  colbacktitle=PromptFrame,
  coltitle=white,
  fonttitle=\sffamily\bfseries\small,
  fontupper=\fontsize{9}{10.5}\selectfont,
  boxrule=0.45pt,
  arc=1.5pt,
  left=9pt,right=9pt,top=5pt,bottom=7pt,
  toptitle=4pt,bottomtitle=4pt,
  before skip=0pt,after skip=8pt,
  before upper={\raggedright\setlength{\parindent}{0pt}\setlength{\parskip}{0pt}}
}
\newcommand{\PromptRole}[1]{%
  \par\addvspace{5pt}%
  {\sffamily\bfseries\fontsize{8}{9}\selectfont\color{PromptFrame}\MakeUppercase{#1}\par}%
  \nobreak\vspace{2pt}%
}
\newcommand{\PromptVar}[1]{\textcolor{PromptVariable}{\texttt{\bfseries\{\detokenize{#1}\}}}}
\newcommand{\PromptToken}[1]{\textcolor{PromptSyntax}{\texttt{\bfseries #1}}}
\newcommand{\PromptGap}{\par\vspace{4pt}}

\usepackage{placeins}

\newcommand{\ScoreWithDelta}[2]{#1\rlap{\,\raisebox{0.55ex}{\scriptsize%
  \if\relax\detokenize{#2}\relax\else
    \ifdim#2pt>0pt\color{ForestGreen}\else
      \ifdim#2pt<0pt\color{BrickRed}\fi
    \fi
    $#2$%
  \fi}}}

\newcommand{\CompactScoreWithDelta}[2]{#1\rlap{\kern0.7pt\raisebox{0.55ex}{\tiny%
  \if\relax\detokenize{#2}\relax\else
    \ifdim#2pt>0pt\color{ForestGreen}\else
      \ifdim#2pt<0pt\color{BrickRed}\fi
    \fi
    \textnormal{#2}%
  \fi}}}

\newtheorem*{thmidasmp*}{Identifying assumptions}

\theoremstyle{definition}

\newtheorem*{thmrem*}{Remark}
\newtheorem*{thmprop*}{Proposition}

\def\bm{{\bf m}}

\iclrfinalcopy 
\begin{document}

\maketitle
\lhead{}
\chead{}
\rhead{}

\begin{abstract}

We study how to build more capable general-purpose agents by
extending large language models (LLMs) with native typed
decision-making.
We introduce \textbf{Dyad}, an architecture that augments a
pretrained LLM with an environment-conditioned action encoder that embeds each candidate action description in parallel, then scores these embeddings against the LLM's internal state to yield a distribution over typed actions.
%
%
By factorizing decision-making into representations of the
evolving interaction state and environment-specific action
semantics, Dyad introduces an inductive bias for learning
reusable representations while keeping action scoring efficient even as the action space grows.
%
We investigate two complementary reinforcement learning settings driven by environment interaction.
%
With the LLM frozen, training the action encoder alone achieves 
consistent gains across four unseen environments, enabling modular
adaptation without modifying any LLM parameters.
Jointly optimizing both components outperforms conventional RL post-training across diverse interactive tasks and model scales, including a 3.80\% average absolute gain on ALFWorld with a 9B model, while improving general knowledge, reasoning, and coding.
\end{abstract}
\section{Introduction}
Solving tasks through interaction with an environment requires two fundamentally different forms of computation~\citep{DBLP:conf/iclr/YaoZYDSN023}: deliberate reasoning about goals or plans, and typed decision-making over
available actions, analogous to Kahneman's System 2 and System 1, respectively~\citep{fast_and_slow_book}. For example, a shopping agent must reason about a customer's requirements and compare products across web pages (System 2), while repeatedly selecting from the available actions in the active page to search, inspect products, choose options, and complete a purchase (System 1).

\begin{figure}[t]
    \centering
    \includegraphics[width=1\linewidth]{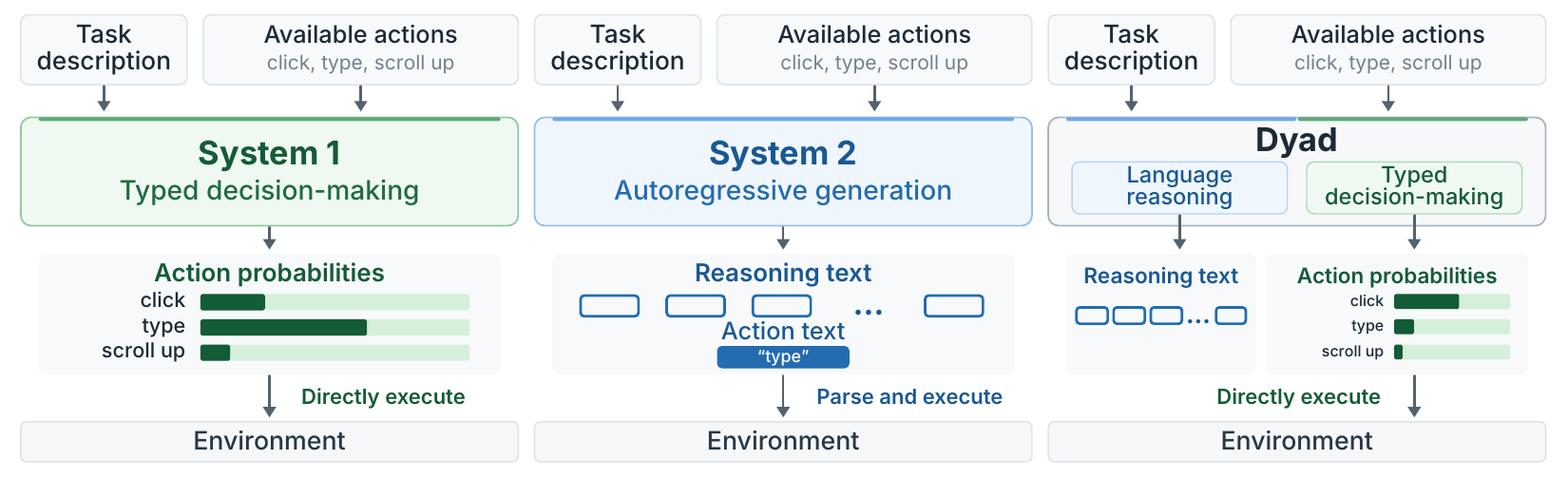}
    \caption{Conceptual comparison of System 1, System 2, and Dyad for reasoning and acting.}
    \vspace{-6mm}
    \label{fig:dyad-overview}
\end{figure}

Today, LLM-based agents typically handle both forms through
autoregressive generation, \ie, reasoning in language and generating text to specify actions, as illustrated in the middle of Figure~\ref{fig:dyad-overview}.
However, this overloads the language-generation interface:
the LLM assigns probabilities to vocabulary tokens rather than
directly to permissible typed actions~\citep{expa}.
Recent dedicated System 1 models, such as Jev~\citep{jev} and its open-source alternatives~\citep{laya,simplejev},
address this limitation by predicting typed decisions directly
and in parallel, rather than generating their textual
representations token by token, as shown on the left of Figure~\ref{fig:dyad-overview}.

Yet language reasoning and typed decision-making need not be
developed independently, as systems 1 and 2 are complementary modes of cognition rather than distinct physical systems
~\citep{fast_and_slow_book}: reasoning guides decisions, while
decision outcomes inform subsequent reasoning.
Recent works~\citep{expa,zhao2026nvcot} take a step toward this integration by extending
LLMs with direct environment actions, enabling interleaved
language reasoning and decision-making.
However, they are limited to fixed, pre-defined action spaces, and do not generalize to dynamically provided actions in unseen environments (\eg, unseen web pages).
This raises a natural question: can pretrained LLMs support native
typed decision-making over dynamic action spaces, and can jointly
learning reasoning and decision-making yield more capable agents?

We introduce \textbf{Dyad}, an architecture that extends pretrained LLMs with \emph{native typed decision-making} while retaining their autoregressive language reasoning capabilities, as illustrated on the right of Figure~\ref{fig:dyad-overview}.
Specifically, we augment the pretrained LLM with an action encoder that embeds available action descriptions independently and in parallel,
conditioned on the environment.
Because these action representations depend on the environment and action semantics, but not on the evolving interaction state, they can be precomputed and reused across decision steps whenever the action definitions remain unchanged.
When a decision is required, these embeddings are scored against
the LLM's internal state representation to produce a distribution
directly over the available typed actions.
This factorization separates the evolving interaction state from
environment-specific action semantics, providing an inductive bias
for learning reusable representations rather than an entangled
state--action mapping.
It further amortizes action encoding across interactions, enabling efficient decision-time scoring and adaptation to previously unseen action spaces.

Dyad is trained in two complementary settings, both driven by end-to-end rewards from environment interaction. First, we freeze the pretrained LLM and train only the action encoder through reinforcement learning. The resulting agent consistently improves performance across four unseen environments, demonstrating that native typed decision-making can enhance existing LLMs without modifying their pretrained parameters. Second, we jointly optimize the LLM and action encoder on diverse interactive tasks. Dyad outperforms agentic RL baselines across model scales. On ALFWorld, it achieves a 3.80\% average absolute gain with Qwen3.5-9B. Dyad also better preserves general knowledge, reasoning, and coding capabilities.

In summary, our contributions are threefold:
\begin{itemize}[itemsep=0pt, left=0pt]
    \item We introduce Dyad, to our knowledge the \emph{first} architecture to extend pretrained LLMs with native, parallel typed decision-making over dynamically provided action spaces.
    
    \item We develop an end-to-end inference and training pipeline integrated with vLLM~\citep{kwon2023vllm} and verl~\citep{sheng2025hybridflow}, which we plan to open-source.
    
    \item We show that Dyad improves agent performance across both frozen-LLM adaptation and joint optimization, generalizes across environments and model scales, and yields broad gains in general knowledge, reasoning, and coding.
\end{itemize}
\section{Dyad: Native Typed Decision-Making in Pretrained LLMs}
\label{sec:2}

\subsection{Problem Formulation}
\label{sec:2.1}

\paragraph{Setting.}
We consider an LLM agent that interleaves language generation with direct interaction in an external environment $e$.
At step $t$, the environment exposes a set of permissible typed actions $\mathcal A_t$, which may change over time.
Each $a\in\mathcal A_t$ is an environment-defined operation,
potentially with arguments, that can be executed directly.
This setting encompasses both direct decisions, such as routing a support request to a team, and tasks combining language reasoning with action, such as navigating unfamiliar software.

\vspace{-2mm}

\paragraph{Policy.}
The agent observes a language history $h_t$, comprising previously generated tokens and textual observations from its interactions.
Its policy $\pi_\theta$ supports two modes:
language generation, with a distribution
$\pi_\theta^{\mathrm{LM}}(v\mid h_t,e)$ over the vocabulary
$\mathcal V$, and typed decision-making, with a distribution
$\pi_\theta^{\mathrm{act}}(a\mid h_t,e,\mathcal A_t)$ over the
currently permissible actions $\mathcal A_t$.

\vspace{-2mm}

\paragraph{Interaction.}
Generating a token $v\in\mathcal V$ extends the language history, whereas selecting a typed action $a\in\mathcal A_t$ invokes the corresponding environment operation.
The resulting textual observation is appended to the history;
for example, an \texttt{open(fridge)} action may return a
description of the fridge's contents.
The environment provides rewards based on task outcomes, such as successfully retrieving a requested item from the fridge.
The agent's objective is to maximize the expected cumulative reward from environment interaction.

To realize such a policy, existing approaches often jointly encode the interaction state and available actions, entangling action semantics with the evolving state~\citep{laya,semif}.
We introduce Dyad, which instead factorizes typed decision-making
into pretrained LLM state representations within a single architecture
(Section~\ref{sec:2.2}), followed by its learning procedure
(Section~\ref{sec:3}).

\subsection{Architecture}
\label{sec:2.2}

\begin{figure}[!t]
    \centering
    \includegraphics[
        width=\linewidth,
        trim={7.96bp 173.56bp 117.88bp 44.64bp}, 
        clip
    ]{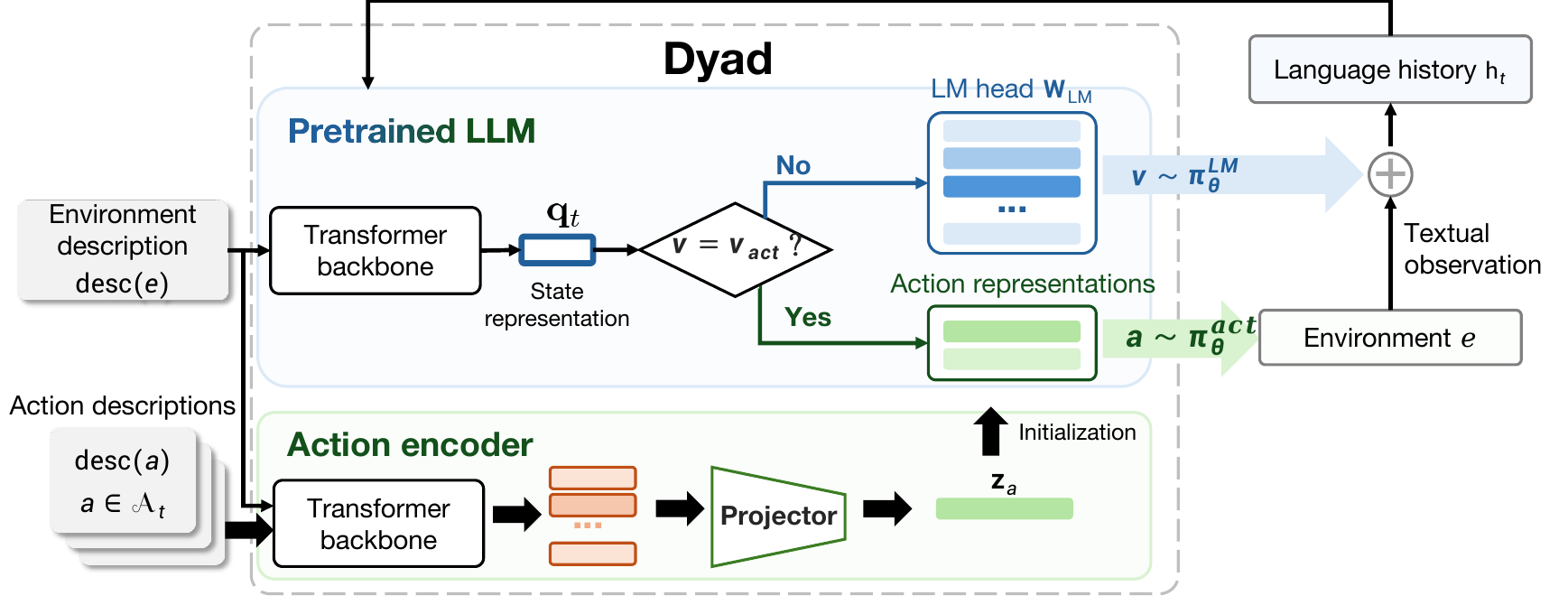}
    \caption{\textbf{Dyad architecture.} State and environment-conditioned action representations are computed independently. Generating the interaction token $v_{\mathrm{act}}$ triggers typed action selection; otherwise, the pretrained LM head generates language tokens.}
    \vspace{-5mm}
    \label{fig:dyad-architecture}
\end{figure}

Dyad extends a pretrained LLM with an action encoder, enabling
language generation and typed decision-making within a single policy.
Let $f$ denote the pretrained LLM's transformer backbone,
$\mathbf W_{\mathrm{LM}}$ its language modeling head, and $g$
the action encoder.
Given the interaction history $h_t$ and environment description
$\mathrm{desc}(e)$, the two components produce a state representation
$\mathbf q_t$ and an environment-conditioned representation
$\mathbf z_a$ for each available action $a\in\mathcal A_t$:
\begin{equation}
    \mathbf q_t=f(\mathrm{desc}(e)\oplus h_t),
\qquad
\mathbf z_a=g(\mathrm{desc}(e)\oplus\mathrm{desc}(a)),
\label{eq:factorize}
\end{equation}

where $\mathrm{desc}(\cdot)$ denotes textual description and $\oplus$ denotes concatenation.
Dyad retains the pretrained language modeling head to generate
tokens according to
\begin{equation}
\pi_\theta^{\mathrm{LM}}(v\mid h_t,e)
=
\operatorname{softmax}
\bigl(\mathbf W_{\mathrm{LM}}\mathbf q_t\bigr)_v.
\label{eq:lm}
\end{equation}
Upon generating a dedicated interaction token
$v_{\mathrm{act}}\in\mathcal V$, Dyad selects a typed action
by scoring each available action against the state representation:
\begin{equation}
\pi_\theta^{\mathrm{act}}(a\mid h_t,e,\mathcal A_t)
=
\operatorname{softmax}
\bigl(\mathbf Z_t\mathbf q_t\bigr)_a,
\qquad
\mathbf Z_t =
\begin{bmatrix}
\mathbf z_{a_1} & \cdots &
\mathbf z_{a_{|\mathcal A_t|}}
\end{bmatrix}^{\top}.
\label{eq:act}
\end{equation}
The key design choice in Dyad is to factorize the computation of
$\pi_\theta^{\mathrm{act}}(a\mid h_t,e,\mathcal A_t)$ into
a state representation $\mathbf q_t$ and an environment-conditioned
action representation $\mathbf z_a$ with Eq.~\ref{eq:factorize}.
This encodes the inductive bias that the evolving interaction state
can be represented independently of the currently available actions,
while an action's meaning depends on its environment rather than
the current state.
The action encoder therefore learns to express action semantics in the pretrained LLM's representation space, enabling their compatibility to be measured directly.
Moreover, independent action encoding makes the decision policy
permutation-equivariant: reordering the available actions simply
reorders their probabilities.
We detail the two representations below.

\noindent\textbf{State Representation.}
We take the pretrained LLM's hidden state at the current decision
position as $\mathbf q_t$.
Pretrained for next-token prediction, the backbone provides contextual
features that support both language generation and typed decision-making.

\noindent\textbf{Action Representation.}
We implement $g$ as a transformer that maps the concatenated
environment and action descriptions to an embedding compatible
with $\mathbf q_t$.
Action embeddings are computed independently and in parallel,
and can be reused across decision steps as long as their
descriptions and environment context remain unchanged.

\noindent\textbf{Amortized Action Encoding.} Action representations can be computed once and reused whenever their action descriptions and environment context remain unchanged. Newly introduced actions need only be encoded when first encountered. Consequently, steady-state decision-time inference requires only the pretrained LLM and cached action representations.
\section{Learning Dyad from Environment Interaction}
\label{sec:3}

We first initialize the action encoder to align its
representations $\mathbf z_a$ with the LLM's state
representations $\mathbf q_t$, then optimize Dyad through
reinforcement learning (RL) from environment interaction.

\subsection{Action Encoder Initialization}
\label{sec:3.1}

The action encoder $g$ consists of a pretrained transformer backbone,
separate from $f$, and a projector that maps its outputs to action
representations $\mathbf z_a$ with the same dimensionality as
$\mathbf q_t$.
To initialize the projector, we synthesize a small collection of
action-selection examples
$\mathcal D=\{(h_i,e_i,\mathcal A_i,a_i^\star)\}_{i=1}^N$.
For each example, we sample a permissible action set
$\mathcal A_i$ for environment $e_i$ and select a target action
$a_i^\star\in\mathcal A_i$.
A teacher LLM then generates task context and reasoning $h_i$
that support selecting $a_i^\star$.
We initialize the projector by minimizing the cross-entropy loss,
while keeping $f$, $\mathbf W_{\mathrm{LM}}$, and the action
encoder backbone frozen:
\begin{equation}
\label{eq:dyad-alignment}
\mathcal L_{\mathrm{align}}
=
-\mathbb E_{(h,e,\mathcal A,a^\star)\sim\mathcal D}
\left[
\log\pi_\theta^{\mathrm{act}}
\bigl(a^\star\mid h,e,\mathcal A\bigr)
\right].
\end{equation}
This lightweight initialization aligns the two representation
spaces before learning from environment rewards.
Further details are provided in
Appendix~\ref{app:action_alignment}.

\subsection{Agentic Reinforcement Learning}
\label{sec:3.2}

Starting from the initialized model, we optimize Dyad using
rewards from environment interaction.
Trajectories interleave language generation and typed
decision-making under the policy defined in
Eqs.~\ref{eq:lm} and~\ref{eq:act}.
We apply GRPO or GiGPO using task-outcome rewards, treating
sampled vocabulary tokens and typed actions as decisions
under their respective policy distributions.
We consider two training settings.
In joint optimization, we update both the pretrained LLM and
the entire action encoder.
In frozen-LLM adaptation, we keep $f$ and
$\mathbf W_{\mathrm{LM}}$ fixed and update only $g$,
including its backbone and projector.
Further optimization details are provided in
Appendix~\ref{app:agentic_rl}.
\section{Related Work}

\paragraph{Language-Based Action Interfaces.}
LLM agents commonly interact with external environments by
generating textual action specifications, as in reasoning and
tool-use agents~\citep{
DBLP:conf/iclr/YaoZYDSN023,
DBLP:conf/nips/SchickDDRLHZCS23}.
Program-guided methods use generated code to invoke external
computation~\citep{
DBLP:conf/icml/GaoMZ00YCN23,
DBLP:conf/iclr/GouSGSYHDC24},
while interactive agents extend this interface to web navigation
and multi-step environment control~\citep{
zhou2023webarena,
liu2023agentbench}.
These approaches offer a flexible interface, but express
environment actions through vocabulary-token prediction
rather than a policy directly over permissible typed actions.

\vspace{-3mm}

\paragraph{Typed Decision-Making.}
Dedicated decision models such as Jev~\citep{jev} predict
distributions directly over provided choices.
As summarized in Table~\ref{tab:decision_parameterizations},
such models admit several architectural parameterizations.
Joint action-set encoders, as in Laya~\citep{laya}, process the
interaction state and available actions together, producing
decision scores for all actions in parallel.
AR encoders instead process each state--action pair to assign
a scalar score, as exemplified by the reranker-based approach
in SemIf~\citep{semif}.
Single-token AR methods assign vocabulary identifiers to
available choices and score them using the LLM's language
modeling head, as in Simple Jev~\citep{simplejev}.
However, these approaches entangle action scoring with the evolving interaction state or a fixed vocabulary basis.

\begin{table}[t]
    \centering
    \caption{Conceptual comparison of methods for typed decision-making. $F$ denotes a model jointly processing its specified inputs, and $\operatorname{id}(a)$ maps an available action to a vocabulary identifier.}
    \label{tab:decision_parameterizations}

    \begin{tabular}{@{}lcl@{}}
        \toprule
        Approach & Logit for $a\in\mathcal A_t$ & Action interface \\
        \midrule
        Joint action-set encoder
        & $F_a(e,h_t,\mathcal A_t)$
        & Joint scoring \\
        \addlinespace
        AR encoder
        & $F(e,h_t,a)$
        & Independent scoring \\
        \addlinespace
        AR Single-token
        & $\mathbf w_{\operatorname{id}(a)}^\top F(e,h_t,\mathcal A_t)$
        & Fixed token identifiers \\
        \addlinespace
        Dyad
        & $f(e,h_t)^\top g(e,a)$
        & Compositional scoring \\
        \bottomrule
    \end{tabular}
    \vspace{-2mm}
\end{table}

\vspace{-3mm}

\paragraph{Native Actions in Pretrained LLMs.}
Beyond dedicated decision models, recent work integrates
language reasoning and environment actions within pretrained LLMs.
ToolkenGPT~\citep{DBLP:conf/nips/HaoLWH23} introduces learned tool embeddings into the language modeling head, enabling tool invocation through autoregressive generation of special tokens.
ExpA~\citep{expa} goes beyond the vocabulary-token interface by
extending LLMs with directly executable actions, enabling interleaved language reasoning and native decision-making.
However, both rely on predefined action spaces.
Dyad extends this direction by factorizing typed decision-making
into pretrained LLM state representations and independently encoded,
environment-conditioned action representations, constructing a native decision head over dynamically provided actions.
\section{Experiments}
\label{sec:experiments}

\subsection{Experimental Setup}
\label{sec:experimental_setup}

\noindent\textbf{Benchmarks.}
We perform agentic RL separately on DIVE~\citep{chen2026dive} and
CodeGym~\citep{du2025codegym}, and evaluate the resulting agents on
ALFWorld~\citep{shridhar2020alfworld}, WebShop~\citep{yao2022webshop},
$\tau^2$-bench~\citep{barres2025tau2bench}, and SWE-bench Verified (SWE)~\citep{jimenez2023swebench} without further training on these environments. We assess general knowledge, mathematical reasoning, and coding with MMLU-Pro~\citep{wang2024mmlupro},
\href{https://huggingface.co/datasets/MathArena/hmmt_feb_2026}{HMMT February 2026}, and LiveCodeBench v6~\citep{jain2024livecodebench}, respectively.
We also train agents separately on ALFWorld and WebShop to evaluate
Dyad's in-domain gains over the corresponding RL baselines.

\noindent\textbf{Baselines.}
We compare Dyad-GRPO with GRPO~\citep{shao2024deepseekmath} and
Dyad-GiGPO with GiGPO~\citep{feng2025gigpo}, using the same LLM and
RL training environment for each pair.
We also evaluate each model without additional training as a reference for agent performance and general capabilities.

\noindent\textbf{Implementation Details.}
For joint training, we use Qwen3.5-4B, Qwen3.5-9B, and Qwen3.5-27B~\citep{qwen2026qwen35}. For frozen-LLM adaptation, we use the 4B and 9B Qwen models, Llama-3.2-3B-Instruct~\citep{meta2024llama32}, and Gemma-4-E2B-it~\citep{gemmateam2026gemma4}. Dyad uses attention pooling as the default projector and follows the alignment and agentic RL procedure in Section~\ref{sec:3}.
We implement the training and inference pipeline with
verl~\citep{sheng2025hybridflow} and vLLM~\citep{kwon2023vllm}.
Further implementation details are provided in Appendices~\ref{app:implementation_details} and~\ref{app:experimental_details}.

\subsection{Agentic Post-training}
\label{sec:agentic_effectiveness}

\begin{table}[t]
    \centering
    \captionsetup{skip=2.0pt}
    \caption{\textbf{In-domain performance of Qwen3.5-9B.} Evaluations use 3 seeds.}
    \label{tab:task-specific-performance}
    \setlength{\tabcolsep}{4pt}
    \renewcommand{\arraystretch}{0.95}
    \begin{tabular}{lcc}
        \toprule
        Method & ALFWorld & WebShop \\
        \midrule
        Base model & 45.66 & 33.43 \\
        \addlinespace[0.3pt]
        \quad + GRPO & 85.01 & 73.44 \\
        \quad + Dyad-GRPO & \textbf{90.90} & \textbf{75.17} \\
        \addlinespace[0.3pt]
        \quad + GiGPO & 89.32 & 81.52 \\
        \quad + Dyad-GiGPO & \textbf{91.02} & \textbf{83.65} \\
        \bottomrule
    \end{tabular}
\end{table}

\noindent\textbf{In-Domain Performance.} When training and evaluation use the same environment, Dyad improves both GRPO and GiGPO on Qwen3.5-9B across ALFWorld and WebShop, outperforming the corresponding baseline in all four comparisons (Table~\ref{tab:task-specific-performance}). The largest absolute gain is 5.89\% on ALFWorld with GRPO. GiGPO already outperforms GRPO in both environments, and Dyad provides further improvements over this stronger baseline.

\noindent\textbf{Cross-Environment Performance.}
When trained on DIVE or CodeGym and evaluated without further
adaptation, Dyad outperforms the corresponding RL baseline in
41 of 48 model--algorithm--environment comparisons
(Table~\ref{tab:joint-generalization}).
It also improves over the original pretrained model in 47 of 48
evaluations, with gains spanning both training environments and all
three model scales.
These results show that the benefits of Dyad are not confined to the environment used for agentic RL.


\begin{table}[!t]
    \caption{\textbf{Cross-environment performance.} Agents are evaluated without further training. Superscripts show absolute changes (\%) from the corresponding RL baseline. ALF denotes ALFWorld. Evaluations use 3 seeds.}
    \label{tab:joint-generalization}
    \vspace{-1mm}
    \begin{center}
    \setlength{\tabcolsep}{0pt}
    \renewcommand{\arraystretch}{1.00}
    \begin{tabular*}{\linewidth}{@{\extracolsep{\fill}}l*{4}{c}@{\hspace{2pt}}*{4}{c}@{}}
        \toprule
         & \multicolumn{4}{c}{\textbf{DIVE}} & \multicolumn{4}{c}{\textbf{CodeGym}} \\
        \cmidrule(lr){2-5}\cmidrule(l){6-9}
        Method & ALF & WebShop & $\tau^2$ & SWE & ALF & WebShop & $\tau^2$ & SWE \\
        \midrule
        \textit{Qwen3.5-4B} & \makebox[22.5pt][r]{23.45}\makebox[14.5pt][l]{} & \makebox[22.5pt][r]{28.54}\makebox[14.5pt][l]{} & \makebox[22.5pt][r]{61.05}\makebox[14.5pt][l]{} & \makebox[22.5pt][r]{43.21}\makebox[14.5pt][l]{} & \makebox[22.5pt][r]{23.45}\makebox[14.5pt][l]{} & \makebox[22.5pt][r]{28.54}\makebox[14.5pt][l]{} & \makebox[22.5pt][r]{61.05}\makebox[14.5pt][l]{} & \makebox[22.5pt][r]{43.21}\makebox[14.5pt][l]{} \\
        \addlinespace[2.5pt]
        +\ GRPO & \makebox[22.5pt][r]{29.71}\makebox[14.5pt][l]{} & \makebox[22.5pt][r]{30.42}\makebox[14.5pt][l]{} & \makebox[22.5pt][r]{60.52}\makebox[14.5pt][l]{} & \makebox[22.5pt][r]{\textbf{45.67}}\makebox[14.5pt][l]{} & \makebox[22.5pt][r]{32.18}\makebox[14.5pt][l]{} & \makebox[22.5pt][r]{33.65}\makebox[14.5pt][l]{} & \makebox[22.5pt][r]{64.10}\makebox[14.5pt][l]{} & \makebox[22.5pt][r]{\textbf{47.98}}\makebox[14.5pt][l]{} \\
        +\ Dyad-GRPO & \makebox[22.5pt][r]{\textbf{31.52}}\makebox[14.5pt][l]{\textsuperscript{\normalfont\fontsize{6}{7}\selectfont\textcolor{ForestGreen}{+1.81}}} & \makebox[22.5pt][r]{\textbf{31.22}}\makebox[14.5pt][l]{\textsuperscript{\normalfont\fontsize{6}{7}\selectfont\textcolor{ForestGreen}{+0.80}}} & \makebox[22.5pt][r]{\textbf{64.54}}\makebox[14.5pt][l]{\textsuperscript{\normalfont\fontsize{6}{7}\selectfont\textcolor{ForestGreen}{+4.02}}} & \makebox[22.5pt][r]{44.28}\makebox[14.5pt][l]{\textsuperscript{\normalfont\fontsize{6}{7}\selectfont\textcolor{BrickRed}{\textminus1.39}}} & \makebox[22.5pt][r]{\textbf{33.65}}\makebox[14.5pt][l]{\textsuperscript{\normalfont\fontsize{6}{7}\selectfont\textcolor{ForestGreen}{+1.47}}} & \makebox[22.5pt][r]{\textbf{35.71}}\makebox[14.5pt][l]{\textsuperscript{\normalfont\fontsize{6}{7}\selectfont\textcolor{ForestGreen}{+2.06}}} & \makebox[22.5pt][r]{\textbf{65.26}}\makebox[14.5pt][l]{\textsuperscript{\normalfont\fontsize{6}{7}\selectfont\textcolor{ForestGreen}{+1.16}}} & \makebox[22.5pt][r]{45.02}\makebox[14.5pt][l]{\textsuperscript{\normalfont\fontsize{6}{7}\selectfont\textcolor{BrickRed}{\textminus2.96}}} \\
        \addlinespace[2.5pt]
        +\ GiGPO & \makebox[22.5pt][r]{28.22}\makebox[14.5pt][l]{} & \makebox[22.5pt][r]{29.54}\makebox[14.5pt][l]{} & \makebox[22.5pt][r]{61.84}\makebox[14.5pt][l]{} & \makebox[22.5pt][r]{45.02}\makebox[14.5pt][l]{} & \makebox[22.5pt][r]{31.09}\makebox[14.5pt][l]{} & \makebox[22.5pt][r]{27.97}\makebox[14.5pt][l]{} & \makebox[22.5pt][r]{63.48}\makebox[14.5pt][l]{} & \makebox[22.5pt][r]{46.64}\makebox[14.5pt][l]{} \\
        +\ Dyad-GiGPO & \makebox[22.5pt][r]{\textbf{33.07}}\makebox[14.5pt][l]{\textsuperscript{\normalfont\fontsize{6}{7}\selectfont\textcolor{ForestGreen}{+4.85}}} & \makebox[22.5pt][r]{\textbf{32.45}}\makebox[14.5pt][l]{\textsuperscript{\normalfont\fontsize{6}{7}\selectfont\textcolor{ForestGreen}{+2.91}}} & \makebox[22.5pt][r]{\textbf{61.86}}\makebox[14.5pt][l]{\textsuperscript{\normalfont\fontsize{6}{7}\selectfont\textcolor{ForestGreen}{+0.02}}} & \makebox[22.5pt][r]{\textbf{47.10}}\makebox[14.5pt][l]{\textsuperscript{\normalfont\fontsize{6}{7}\selectfont\textcolor{ForestGreen}{+2.08}}} & \makebox[22.5pt][r]{\textbf{31.72}}\makebox[14.5pt][l]{\textsuperscript{\normalfont\fontsize{6}{7}\selectfont\textcolor{ForestGreen}{+0.63}}} & \makebox[22.5pt][r]{\textbf{31.21}}\makebox[14.5pt][l]{\textsuperscript{\normalfont\fontsize{6}{7}\selectfont\textcolor{ForestGreen}{+3.24}}} & \makebox[22.5pt][r]{\textbf{64.78}}\makebox[14.5pt][l]{\textsuperscript{\normalfont\fontsize{6}{7}\selectfont\textcolor{ForestGreen}{+1.30}}} & \makebox[22.5pt][r]{\textbf{49.61}}\makebox[14.5pt][l]{\textsuperscript{\normalfont\fontsize{6}{7}\selectfont\textcolor{ForestGreen}{+2.97}}} \\
        \midrule
        \textit{Qwen3.5-9B} & \makebox[22.5pt][r]{45.66}\makebox[14.5pt][l]{} & \makebox[22.5pt][r]{33.43}\makebox[14.5pt][l]{} & \makebox[22.5pt][r]{62.17}\makebox[14.5pt][l]{} & \makebox[22.5pt][r]{44.56}\makebox[14.5pt][l]{} & \makebox[22.5pt][r]{45.66}\makebox[14.5pt][l]{} & \makebox[22.5pt][r]{33.43}\makebox[14.5pt][l]{} & \makebox[22.5pt][r]{62.17}\makebox[14.5pt][l]{} & \makebox[22.5pt][r]{44.56}\makebox[14.5pt][l]{} \\
        \addlinespace[2.5pt]
        +\ GRPO & \makebox[22.5pt][r]{52.67}\makebox[14.5pt][l]{} & \makebox[22.5pt][r]{34.52}\makebox[14.5pt][l]{} & \makebox[22.5pt][r]{62.06}\makebox[14.5pt][l]{} & \makebox[22.5pt][r]{\textbf{47.22}}\makebox[14.5pt][l]{} & \makebox[22.5pt][r]{51.55}\makebox[14.5pt][l]{} & \makebox[22.5pt][r]{36.23}\makebox[14.5pt][l]{} & \makebox[22.5pt][r]{61.31}\makebox[14.5pt][l]{} & \makebox[22.5pt][r]{49.40}\makebox[14.5pt][l]{} \\
        +\ Dyad-GRPO & \makebox[22.5pt][r]{\textbf{53.82}}\makebox[14.5pt][l]{\textsuperscript{\normalfont\fontsize{6}{7}\selectfont\textcolor{ForestGreen}{+1.15}}} & \makebox[22.5pt][r]{\textbf{36.31}}\makebox[14.5pt][l]{\textsuperscript{\normalfont\fontsize{6}{7}\selectfont\textcolor{ForestGreen}{+1.79}}} & \makebox[22.5pt][r]{\textbf{64.88}}\makebox[14.5pt][l]{\textsuperscript{\normalfont\fontsize{6}{7}\selectfont\textcolor{ForestGreen}{+2.82}}} & \makebox[22.5pt][r]{46.20}\makebox[14.5pt][l]{\textsuperscript{\normalfont\fontsize{6}{7}\selectfont\textcolor{BrickRed}{\textminus1.02}}} & \makebox[22.5pt][r]{\textbf{55.13}}\makebox[14.5pt][l]{\textsuperscript{\normalfont\fontsize{6}{7}\selectfont\textcolor{ForestGreen}{+3.58}}} & \makebox[22.5pt][r]{\textbf{38.07}}\makebox[14.5pt][l]{\textsuperscript{\normalfont\fontsize{6}{7}\selectfont\textcolor{ForestGreen}{+1.84}}} & \makebox[22.5pt][r]{\textbf{63.02}}\makebox[14.5pt][l]{\textsuperscript{\normalfont\fontsize{6}{7}\selectfont\textcolor{ForestGreen}{+1.71}}} & \makebox[22.5pt][r]{\textbf{50.62}}\makebox[14.5pt][l]{\textsuperscript{\normalfont\fontsize{6}{7}\selectfont\textcolor{ForestGreen}{+1.22}}} \\
        \addlinespace[2.5pt]
        +\ GiGPO & \makebox[22.5pt][r]{50.98}\makebox[14.5pt][l]{} & \makebox[22.5pt][r]{36.23}\makebox[14.5pt][l]{} & \makebox[22.5pt][r]{\textbf{63.52}}\makebox[14.5pt][l]{} & \makebox[22.5pt][r]{48.83}\makebox[14.5pt][l]{} & \makebox[22.5pt][r]{52.08}\makebox[14.5pt][l]{} & \makebox[22.5pt][r]{\textbf{36.58}}\makebox[14.5pt][l]{} & \makebox[22.5pt][r]{64.21}\makebox[14.5pt][l]{} & \makebox[22.5pt][r]{49.78}\makebox[14.5pt][l]{} \\
        +\ Dyad-GiGPO & \makebox[22.5pt][r]{\textbf{52.06}}\makebox[14.5pt][l]{\textsuperscript{\normalfont\fontsize{6}{7}\selectfont\textcolor{ForestGreen}{+1.08}}} & \makebox[22.5pt][r]{\textbf{37.02}}\makebox[14.5pt][l]{\textsuperscript{\normalfont\fontsize{6}{7}\selectfont\textcolor{ForestGreen}{+0.79}}} & \makebox[22.5pt][r]{61.90}\makebox[14.5pt][l]{\textsuperscript{\normalfont\fontsize{6}{7}\selectfont\textcolor{BrickRed}{\textminus1.62}}} & \makebox[22.5pt][r]{\textbf{50.42}}\makebox[14.5pt][l]{\textsuperscript{\normalfont\fontsize{6}{7}\selectfont\textcolor{ForestGreen}{+1.59}}} & \makebox[22.5pt][r]{\textbf{55.01}}\makebox[14.5pt][l]{\textsuperscript{\normalfont\fontsize{6}{7}\selectfont\textcolor{ForestGreen}{+2.93}}} & \makebox[22.5pt][r]{35.10}\makebox[14.5pt][l]{\textsuperscript{\normalfont\fontsize{6}{7}\selectfont\textcolor{BrickRed}{\textminus1.48}}} & \makebox[22.5pt][r]{\textbf{65.85}}\makebox[14.5pt][l]{\textsuperscript{\normalfont\fontsize{6}{7}\selectfont\textcolor{ForestGreen}{+1.64}}} & \makebox[22.5pt][r]{\textbf{51.25}}\makebox[14.5pt][l]{\textsuperscript{\normalfont\fontsize{6}{7}\selectfont\textcolor{ForestGreen}{+1.47}}} \\
        \midrule
        \textit{Qwen3.5-27B} & \makebox[22.5pt][r]{68.45}\makebox[14.5pt][l]{} & \makebox[22.5pt][r]{44.56}\makebox[14.5pt][l]{} & \makebox[22.5pt][r]{68.21}\makebox[14.5pt][l]{} & \makebox[22.5pt][r]{66.78}\makebox[14.5pt][l]{} & \makebox[22.5pt][r]{68.45}\makebox[14.5pt][l]{} & \makebox[22.5pt][r]{44.56}\makebox[14.5pt][l]{} & \makebox[22.5pt][r]{68.21}\makebox[14.5pt][l]{} & \makebox[22.5pt][r]{66.78}\makebox[14.5pt][l]{} \\
        \addlinespace[2.5pt]
        +\ GRPO & \makebox[22.5pt][r]{69.25}\makebox[14.5pt][l]{} & \makebox[22.5pt][r]{46.68}\makebox[14.5pt][l]{} & \makebox[22.5pt][r]{69.82}\makebox[14.5pt][l]{} & \makebox[22.5pt][r]{68.92}\makebox[14.5pt][l]{} & \makebox[22.5pt][r]{\textbf{71.89}}\makebox[14.5pt][l]{} & \makebox[22.5pt][r]{42.13}\makebox[14.5pt][l]{} & \makebox[22.5pt][r]{69.29}\makebox[14.5pt][l]{} & \makebox[22.5pt][r]{67.91}\makebox[14.5pt][l]{} \\
        +\ Dyad-GRPO & \makebox[22.5pt][r]{\textbf{71.42}}\makebox[14.5pt][l]{\textsuperscript{\normalfont\fontsize{6}{7}\selectfont\textcolor{ForestGreen}{+2.17}}} & \makebox[22.5pt][r]{\textbf{48.81}}\makebox[14.5pt][l]{\textsuperscript{\normalfont\fontsize{6}{7}\selectfont\textcolor{ForestGreen}{+2.13}}} & \makebox[22.5pt][r]{\textbf{71.46}}\makebox[14.5pt][l]{\textsuperscript{\normalfont\fontsize{6}{7}\selectfont\textcolor{ForestGreen}{+1.64}}} & \makebox[22.5pt][r]{\textbf{70.24}}\makebox[14.5pt][l]{\textsuperscript{\normalfont\fontsize{6}{7}\selectfont\textcolor{ForestGreen}{+1.32}}} & \makebox[22.5pt][r]{71.11}\makebox[14.5pt][l]{\textsuperscript{\normalfont\fontsize{6}{7}\selectfont\textcolor{BrickRed}{\textminus0.78}}} & \makebox[22.5pt][r]{\textbf{48.29}}\makebox[14.5pt][l]{\textsuperscript{\normalfont\fontsize{6}{7}\selectfont\textcolor{ForestGreen}{+6.16}}} & \makebox[22.5pt][r]{\textbf{70.10}}\makebox[14.5pt][l]{\textsuperscript{\normalfont\fontsize{6}{7}\selectfont\textcolor{ForestGreen}{+0.81}}} & \makebox[22.5pt][r]{\textbf{68.34}}\makebox[14.5pt][l]{\textsuperscript{\normalfont\fontsize{6}{7}\selectfont\textcolor{ForestGreen}{+0.43}}} \\
        \addlinespace[2.5pt]
        +\ GiGPO & \makebox[22.5pt][r]{69.84}\makebox[14.5pt][l]{} & \makebox[22.5pt][r]{\textbf{47.20}}\makebox[14.5pt][l]{} & \makebox[22.5pt][r]{69.56}\makebox[14.5pt][l]{} & \makebox[22.5pt][r]{68.02}\makebox[14.5pt][l]{} & \makebox[22.5pt][r]{70.79}\makebox[14.5pt][l]{} & \makebox[22.5pt][r]{45.91}\makebox[14.5pt][l]{} & \makebox[22.5pt][r]{69.51}\makebox[14.5pt][l]{} & \makebox[22.5pt][r]{67.90}\makebox[14.5pt][l]{} \\
        +\ Dyad-GiGPO & \makebox[22.5pt][r]{\textbf{72.30}}\makebox[14.5pt][l]{\textsuperscript{\normalfont\fontsize{6}{7}\selectfont\textcolor{ForestGreen}{+2.46}}} & \makebox[22.5pt][r]{45.44}\makebox[14.5pt][l]{\textsuperscript{\normalfont\fontsize{6}{7}\selectfont\textcolor{BrickRed}{\textminus1.76}}} & \makebox[22.5pt][r]{\textbf{71.87}}\makebox[14.5pt][l]{\textsuperscript{\normalfont\fontsize{6}{7}\selectfont\textcolor{ForestGreen}{+2.31}}} & \makebox[22.5pt][r]{\textbf{71.53}}\makebox[14.5pt][l]{\textsuperscript{\normalfont\fontsize{6}{7}\selectfont\textcolor{ForestGreen}{+3.51}}} & \makebox[22.5pt][r]{\textbf{73.21}}\makebox[14.5pt][l]{\textsuperscript{\normalfont\fontsize{6}{7}\selectfont\textcolor{ForestGreen}{+2.42}}} & \makebox[22.5pt][r]{\textbf{47.72}}\makebox[14.5pt][l]{\textsuperscript{\normalfont\fontsize{6}{7}\selectfont\textcolor{ForestGreen}{+1.81}}} & \makebox[22.5pt][r]{\textbf{72.30}}\makebox[14.5pt][l]{\textsuperscript{\normalfont\fontsize{6}{7}\selectfont\textcolor{ForestGreen}{+2.79}}} & \makebox[22.5pt][r]{\textbf{68.98}}\makebox[14.5pt][l]{\textsuperscript{\normalfont\fontsize{6}{7}\selectfont\textcolor{ForestGreen}{+1.08}}} \\
        \bottomrule
    \end{tabular*}
    \end{center}
\end{table}

\begin{figure}[t]
    \centering
    \includegraphics[width=\linewidth]{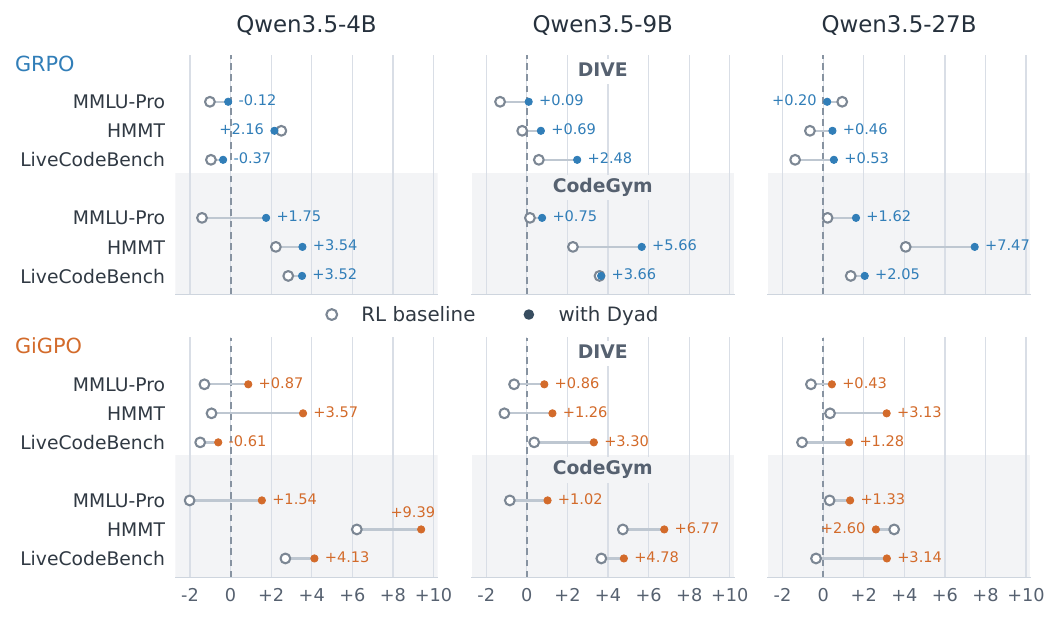}
    \caption{\textbf{General capabilities.} Dashed lines indicate zero change relative to each model without additional training. Evaluations are on 3 seeds.}
    \label{fig:general-capabilities}
\end{figure}

\noindent\textbf{General-Capability Retention.}
Learning both reasoning and environment control through vocabulary
generation can create interference between the two objectives during
agentic RL~\citep{expa}.
We therefore measure changes in general knowledge, mathematical
reasoning, and coding relative to the original pretrained models
(Figure~\ref{fig:general-capabilities}).
Dyad generally outperforms the corresponding GRPO and GiGPO baselines,
both by recovering capability losses after DIVE training and by
amplifying gains after CodeGym training.
Notably, both Dyad variants exceed the original models on every
CodeGym capability evaluation, while the RL baselines still regress
on general knowledge or coding in some settings.
Together with the agent-performance gains, these results suggest that
the action encoder provides an additional route for task adaptation,
reducing interference with pretrained language capabilities.

\begin{figure}[t]
    \centering
    \includegraphics[width=\linewidth]{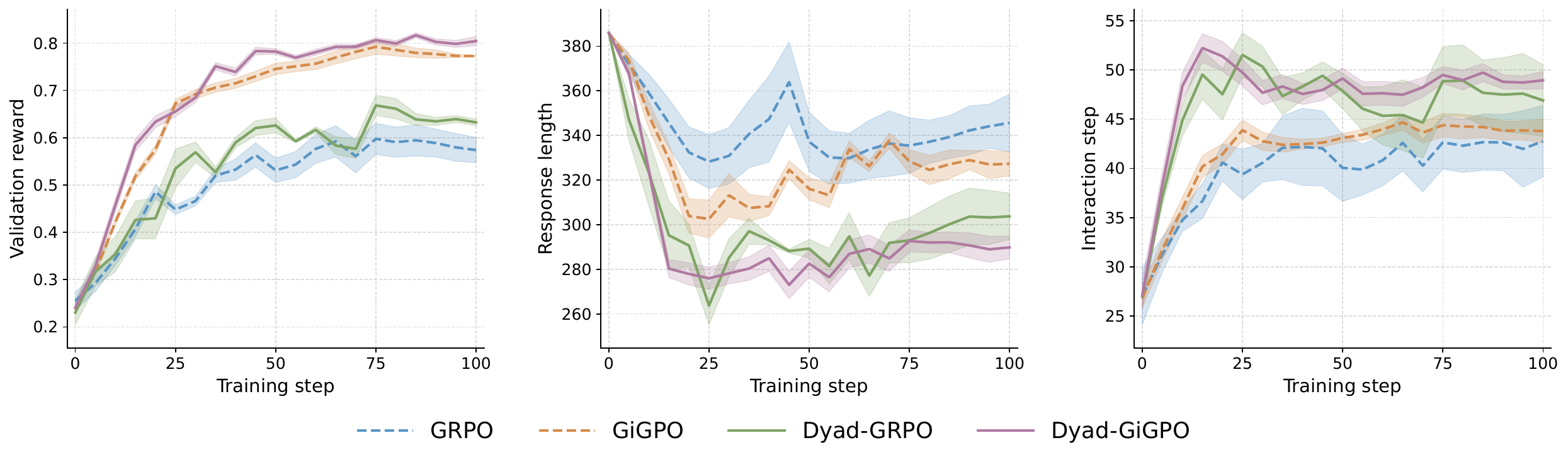}
    \caption{\textbf{Training dynamics.} Curves show validation means for Qwen3.5-9B on CodeGym; shading indicates $\pm1$ standard deviation.}
    \label{fig:codegym-9b-training}
\end{figure}

\noindent\textbf{Training Dynamics.}
Both Dyad variants attain higher final validation rewards than their
corresponding RL baselines (Figure~\ref{fig:codegym-9b-training}).
As training progresses, Dyad produces substantially shorter responses,
suggesting more concise and efficient reasoning, while increasing the
number of environment interactions per trajectory.
This shift toward less generation and more interaction is especially beneficial for multi-step tasks, where progress depends on repeatedly acting on and incorporating feedback from the environment.

\subsection{Modular Adaptation with a Frozen LLM}
\label{sec:environment_adaptation}

\begin{table}[t]
    \centering
    \caption{\textbf{Frozen-LLM adaptation.} Dyad is trained with GRPO on DIVE. Evaluations use 3 seeds.}
    \label{tab:frozen-llm-adaptation}
    \setlength{\tabcolsep}{3pt}
    \renewcommand{\arraystretch}{1.08}
    \begin{tabular*}{\linewidth}{@{\extracolsep{\fill}}l*{4}{w{c}{43pt}}@{\hspace{15pt}}}
        \toprule
        Method & ALFWorld & WebShop & $\tau^2$ & SWE \\
        \midrule
        \textit{Qwen3.5-4B} & \ScoreWithDelta{23.45}{} & \ScoreWithDelta{28.54}{} & \ScoreWithDelta{61.05}{} & \ScoreWithDelta{43.21}{} \\
        \quad + Dyad w/ freezing & \ScoreWithDelta{\textbf{27.75}}{+4.30} & \ScoreWithDelta{\textbf{31.47}}{+2.93} & \ScoreWithDelta{\textbf{63.88}}{+2.83} & \ScoreWithDelta{\textbf{43.78}}{+0.57} \\
        \quad + Dyad w/o freezing & \ScoreWithDelta{31.52}{} & \ScoreWithDelta{31.22}{} & \ScoreWithDelta{64.54}{} & \ScoreWithDelta{44.28}{} \\
        \midrule
        \textit{Qwen3.5-9B} & \ScoreWithDelta{45.66}{} & \ScoreWithDelta{33.43}{} & \ScoreWithDelta{62.17}{} & \ScoreWithDelta{44.56}{} \\
        \quad + Dyad w/ freezing & \ScoreWithDelta{\textbf{48.82}}{+3.16} & \ScoreWithDelta{\textbf{35.00}}{+1.57} & \ScoreWithDelta{\textbf{64.23}}{+2.06} & \ScoreWithDelta{\textbf{45.50}}{+0.94} \\
        \quad + Dyad w/o freezing & \ScoreWithDelta{53.82}{} & \ScoreWithDelta{36.31}{} & \ScoreWithDelta{64.88}{} & \ScoreWithDelta{46.20}{} \\
        \midrule
        \textit{Llama-3.2-3B-Instruct} & \ScoreWithDelta{16.98}{} & \ScoreWithDelta{8.41}{} & \ScoreWithDelta{\textbf{18.72}}{} & \ScoreWithDelta{24.32}{} \\
        \quad + Dyad w/ freezing & \ScoreWithDelta{\textbf{18.32}}{+1.34} & \ScoreWithDelta{\textbf{10.09}}{+1.68} & \ScoreWithDelta{18.44}{-0.28} & \ScoreWithDelta{\textbf{24.80}}{+0.48} \\
        \quad + Dyad w/o freezing & \ScoreWithDelta{21.03}{} & \ScoreWithDelta{12.94}{} & \ScoreWithDelta{20.10}{} & \ScoreWithDelta{25.12}{} \\
        \midrule
        \textit{Gemma-4-E2B-it} & \ScoreWithDelta{24.24}{} & \ScoreWithDelta{17.11}{} & \ScoreWithDelta{31.23}{} & \ScoreWithDelta{\textbf{28.66}}{} \\
        \quad + Dyad w/ freezing & \ScoreWithDelta{\textbf{27.73}}{+3.49} & \ScoreWithDelta{\textbf{19.30}}{+2.19} & \ScoreWithDelta{\textbf{33.94}}{+2.71} & \ScoreWithDelta{28.15}{-0.51} \\
        \quad + Dyad w/o freezing & \ScoreWithDelta{29.95}{} & \ScoreWithDelta{21.48}{} & \ScoreWithDelta{34.01}{} & \ScoreWithDelta{30.56}{} \\
        \bottomrule
    \end{tabular*}
\end{table}

\noindent\textbf{Frozen-LLM Adaptation.}
With the LLM frozen, training the action encoder improves agent
performance over the original model in most evaluated settings
(Table~\ref{tab:frozen-llm-adaptation}).
Both Qwen models improve on all four environments, and the gains extend
to other model families, with improvements on WebShop for every model.
Joint training generally achieves higher scores, but the gap varies by
model and environment: for both Qwen models and Gemma, frozen adaptation
is close on $\tau^2$-bench, whereas larger gaps remain on ALFWorld.
For both Qwen models, the gap on SWE is also small.
These results show that training only the action encoder can improve
agent performance beyond the RL training environment without updating
the LLM.

\subsection{System 1 Evaluation}
\label{sec:in_depth_analysis}

\begin{table}[t]
    \centering
    \caption{\textbf{JevBench results.} Accuracy (\%) on the public subset. Best accuracy and latency (s) values are bold. $\dagger$~Jev latency comes from published API measurements. Benchmark and measurement details are provided in Appendix~\ref{app:jevbench}.}
    \label{tab:jevbench-results}
    \setlength{\tabcolsep}{3.5pt}
    \begin{tabular}{@{}lcccccc@{}}
        \toprule
        Method & Easy & Standard & Hard & Overall & Latency  & Output tokens \\
        \midrule
        Jev 1.13 & \textbf{100.00} & 96.79 & 70.88 & 85.01 & $0.67^{\dagger}$ & --- \\
        Qwen3.5-4B (w/o reasoning) & \textbf{100.00} & 96.41 & 56.32 & 77.89 & 0.32 & --- \\
        Qwen3.5-4B (w/ reasoning) & 98.76 & 95.62 & 59.88 & 79.10 & 1.02 & 102.82 \\
        Qwen3.5-4B + Dyad (w/o reasoning) & \textbf{100.00} & 97.88 & 67.07 & 83.52 & \textbf{0.31} & --- \\
        Qwen3.5-4B + Dyad (w/ reasoning) & \textbf{100.00} & \textbf{98.72} & \textbf{72.08} & \textbf{86.18} & 1.35 & 120.44 \\
        \bottomrule
    \end{tabular}
\end{table}

\noindent\textbf{Comparison with Dedicated System 1 Models.}
We evaluate Qwen3.5-4B + Dyad, trained on CodeGym using GRPO, against Jev on both direct decision-making and agentic interaction.
On JevBench, Dyad with reasoning reaches 86.18\% overall accuracy, slightly exceeding Jev's 85.01\%, with the largest advantage on the hard subset (Table~\ref{tab:jevbench-results}).
Without reasoning, Dyad remains competitive while reducing latency
to 0.31\,s, compared with 0.67\,s reported for Jev.
On the more agentic ALFWorld benchmark, Dyad achieves higher success
rates than both Jev and zero-shot Qwen3.5-4B across all six task types,
while requiring fewer interaction steps overall and on most task types
(Figure~\ref{fig:jev-comparison}).
Together, these results show that Dyad can serve as an effective
typed decision model while retaining autoregressive reasoning within
the same policy.

%
%
%
\begingroup
\setlength{\intextsep}{6pt}
\begin{figure}[t]
    \centering
    \includegraphics[width=0.9\linewidth,trim=0 4bp 0 13bp,clip]{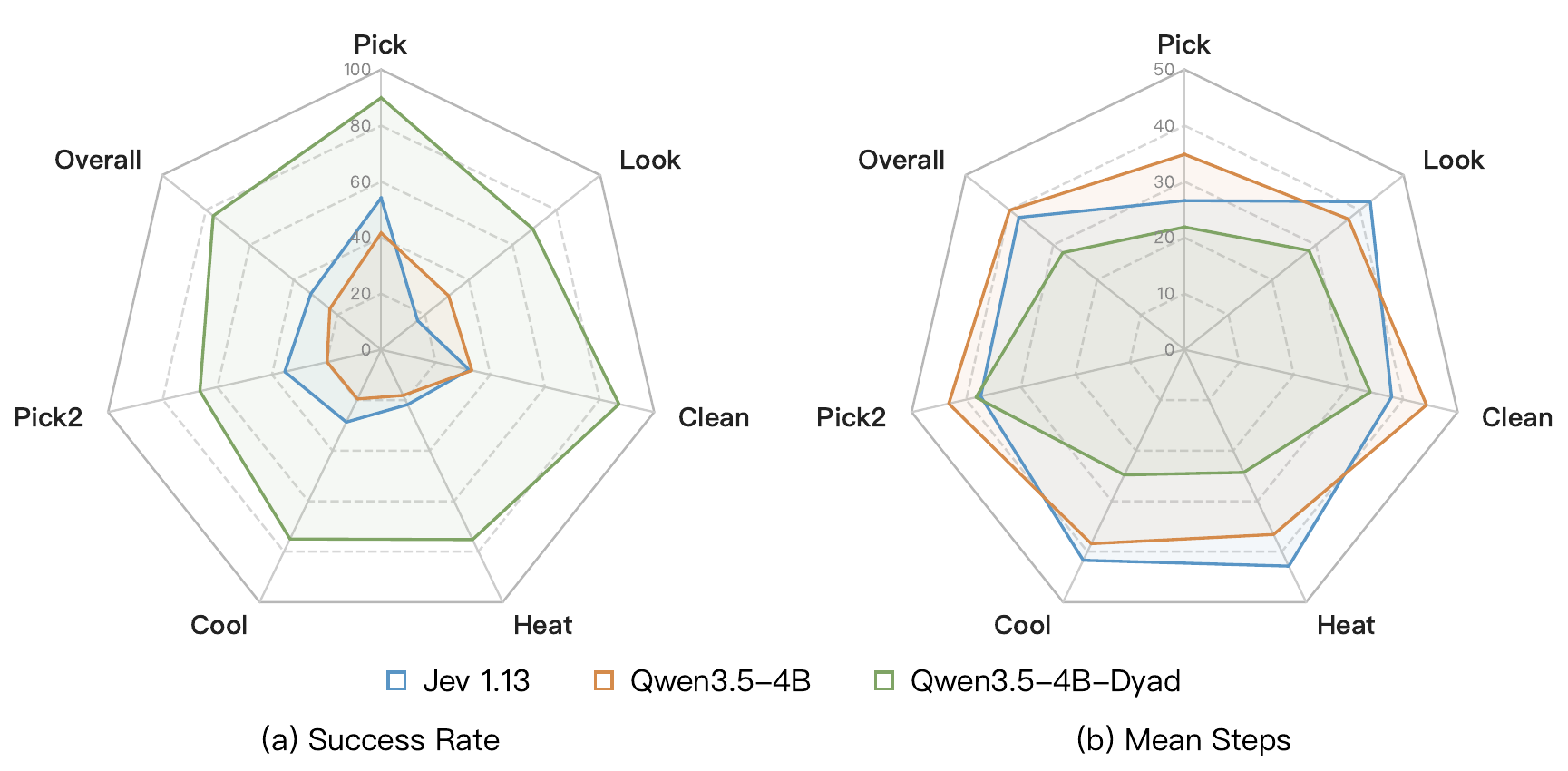}
    \caption{\textbf{Comparison with Jev.} Qwen3.5-4B serves as the zero-shot reference on ALFWorld.}
    \label{fig:jev-comparison}
\end{figure}
\endgroup

\subsection{Ablation Studies}
\label{sec:ablation_analysis}

\begin{figure}[t]
    \centering
    \begingroup
    \phantomsubcaption\label{fig:ablation-components}
    \phantomsubcaption\label{fig:ablation-alignment}
    \endgroup
    \includegraphics[width=\linewidth]{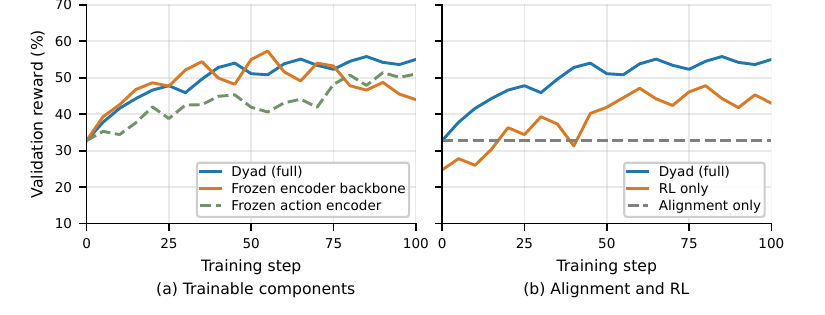}
    \caption{\textbf{Ablation studies.} CodeGym validation reward with Qwen3.5-4B, GRPO, and attention pooling. Both panels reuse the Dyad (full) curve; the horizontal dashed line denotes the alignment-only reference.}
    \label{fig:ablation-training}
\end{figure}

\noindent\textbf{Trainable Components.}
Updating the full action encoder yields the highest final validation reward,
with the LLM trainable in all variants (Figure~\ref{fig:ablation-components}).
Training the projector while freezing the encoder backbone is competitive
early on, but reward declines later and ends below the variant with the
entire encoder frozen. Even with fixed action representations, LLM updates
improve reward over the shared aligned initialization.

\noindent\textbf{Alignment and Reinforcement Learning.}
Alignment improves initial performance, and the full model outperforms
RL-only training throughout the evaluated budget
(Figure~\ref{fig:ablation-alignment}). Although RL without alignment improves
substantially, it does not close the gap. Continuing RL from the aligned
checkpoint also raises reward above the alignment-only reference. These
comparisons support complementary roles for the two stages: supervised
alignment provides a useful initialization, while agentic RL further improves
task performance.

\noindent\begin{wraptable}{r}{0.45\textwidth}
\vspace{4.5mm}
    \centering
    \captionsetup{
        justification=raggedright,
        singlelinecheck=false,
        skip=2pt
    }
    \caption{\textbf{Action pooling ablation.}}
    \label{tab:ablation-projector}
    \setlength{\tabcolsep}{3pt}
    \renewcommand{\arraystretch}{0.9}
    \begin{tabular}{@{}lc@{}}
        \toprule
        Projector & Validation reward \\
        \midrule
        Mean pooling & 41.23 \\
        MLP-weighted pooling & 48.20 \\
        Attention pooling & 55.92 \\
        \bottomrule
    \end{tabular}
\end{wraptable}

\noindent\textbf{Projector Architecture.} Validation reward is highest with attention pooling, followed by
MLP-weighted pooling and mean pooling (Table~\ref{tab:ablation-projector}).
In Dyad, each pooled action vector forms a row of the action-scoring matrix
$\mathbf Z_t$ and is compared directly with the LLM state through a dot
product (Equation~\ref{eq:act}). The projector therefore needs to produce
representations suitable for this comparison from a separate encoder's
token features. MLP-weighted pooling learns token weights,
but its output remains a weighted average of the encoder states.
Attention pooling additionally learns value and output projections,
allowing the projector to transform the features used for action scoring
(Appendix~\ref{app:action_encoding}). This gives it a direct way to adapt
the pooled representation to the LLM's state representation space, in
addition to learning token weights. Its advantage over MLP-weighted pooling
is consistent with a benefit from combining token aggregation and feature
transformation.

\section{Conclusion}
We introduced \textbf{Dyad}, an architecture for extending pretrained
LLMs with native typed decision-making over dynamically provided action
spaces.
Dyad factorizes action selection into a representation of the evolving
interaction state and independently encoded, environment-conditioned
action representations, allowing the same pretrained LLM state to support both autoregressive language reasoning and direct typed
decisions.
This design provides a reusable decision interface without requiring
actions to be represented through the language-model vocabulary.
Across agentic post-training experiments, Dyad improves both in-domain
and cross-environment performance over conventional RL baselines across
model scales.
These gains are accompanied by shorter reasoning traces, more frequent
environment interaction, and generally stronger performance on the
evaluated knowledge, reasoning, and coding benchmarks.
Dyad also supports modular adaptation: training only the action encoder
improves agent performance without modifying the pretrained LLM.
Finally, on dedicated System~1 evaluations, Dyad is competitive with
specialized decision models while retaining language reasoning within
the same policy.
Together, these results suggest that explicitly separating language
reasoning from typed decision-making is a promising foundation for
general-purpose agents that must reason and act across diverse,
previously unseen environments.

\subsection*{AI use statement}
We used generative AI to assist with synthetic data generation and literature search. The authors take responsibility for the final content.


\subsection*{Ethics statement}

This work studies decision-making in language model agents using public benchmarks. We encourage responsible use of the proposed method.



\subsection*{Reproducibility statement}

Implementation details, data construction, experimental settings, hyperparameters, and prompt templates are provided in Appendices A and B to support reproducibility.





\bibliography{iclr2027_conference}
\bibliographystyle{iclr2027_conference}

\clearpage
\appendix
\section{Implementation Details}
\label{app:implementation_details}

\subsection{Action Encoding}
\label{app:action_encoding}
The action encoder $g$ uses a pretrained transformer backbone separate from
$f$, followed by a shared projector. Alignment trains only the projector.
Agentic RL updates both the encoder backbone and projector, including when
the LLM is frozen. The projector pools the encoder's token representations
into an action representation with the same dimensionality as $\mathbf q_t$.

Each candidate is encoded with the environment's action descriptions and
its own definition. These descriptions supply the action context in
$\mathrm{desc}(e)$. The permissible set $\mathcal A_t$ determines
which candidates participate in the action distribution at a decision step.
The encoder input excludes the current task context and reasoning in $h_t$.
Action representations can be reused while their input descriptions and
encoder parameters remain unchanged, and are recomputed when either changes.

With these textual inputs fixed, permuting the rows of $\mathbf Z_t$
permutes the corresponding action probabilities. This property concerns
the order of candidates in the scoring matrix. Reordering the action
descriptions within the encoder or LLM input can change their representations
and is not covered by this property.

\paragraph{Tool Calls.}
Each environment interaction involves a single typed decision over
$\mathcal A_t$. The language modeling head generates reasoning and any
required arguments. The selected action's fixed textual form is appended
to $h_t$, even when it spans multiple tokens. Once complete, the call is
executed and the resulting observation is appended to the history.

\paragraph{Projector Architectures.}
Mean pooling uniformly averages encoder token representations and adds no
trainable pooling parameters; this variant skips alignment.
MLP-weighted pooling uses an MLP to score each
token, then computes a softmax-weighted average of the representations.
Attention pooling uses a learnable query to attend to projected keys and
values, followed by an output projection. Padding tokens are masked in all
three variants.

\paragraph{Action Encoder Input Formats.}
\label{app:action_description_robustness}
We use structured MCP-style action descriptions as the default input to
the action encoder. Table~\ref{tab:action-encoder-input-formats} compares
this format with natural-language descriptions for Qwen3.5-4B with
Dyad-GRPO trained on CodeGym. MCP-style inputs yield absolute gains of
2.87\% on ALFWorld and 4.39\% on WebShop. The
consistent advantage across these two evaluation environments supports
structured descriptions as the encoder's default input format beyond
the environment used for agentic RL.

Natural-language descriptions are substantially shorter in both
environments, with a particularly large length difference on ALFWorld.
The performance advantage of MCP-style inputs therefore comes with
additional description tokens. This comparison favors the structured
input in terms of task success, but does not isolate formatting from
differences in description length and content.

\begin{table}[t]
    \centering
    \caption{\textbf{Action encoder input formats.} Qwen3.5-4B with Dyad-GRPO is trained on CodeGym and evaluated on ALFWorld and WebShop. Length denotes the mean number of description tokens.}
    \label{tab:action-encoder-input-formats}
    \setlength{\tabcolsep}{5pt}
    \renewcommand{\arraystretch}{1.12}
    \begin{tabular}{@{}lcccc@{}}
        \toprule
        & \multicolumn{2}{c}{\textbf{ALFWorld}} & \multicolumn{2}{c}{\textbf{WebShop}} \\
        \cmidrule(lr){2-3}\cmidrule(lr){4-5}
        \textbf{Input format} & \textbf{Length} & \textbf{Success rate} & \textbf{Length} & \textbf{Success rate} \\
        \midrule
        MCP-style (default) & 2665.57 & \textbf{33.88} & 396.50 & \textbf{34.26} \\
        Natural language & 132.64 & 31.01 & 151.50 & 29.87 \\
        \bottomrule
    \end{tabular}
\end{table}

\subsection{Action Encoder Initialization}
\label{app:action_alignment}

\paragraph{Data Construction.}
We construct synthetic examples for action selection across five domains:
spatial manipulation, state change, information, workflow and communication,
and transformation. Each domain contains ten training actions and four
additional actions excluded from training, giving 50 training actions and
20 held-out actions. For each action, we obtain a structured definition
specifying its name, purpose, and arguments, and a corresponding
natural-language description.
We construct synthetic action-selection examples across five domains, with
50 training actions and 20 held-out actions. Each example contains 4--10
permissible actions $\mathcal A_i$ from one domain and a target
$a_i^\star\in\mathcal A_i$. GPT-5.5~\citep{openai2026gpt55} generates
task context and reasoning $h_i$ conditioned on these preselected actions
and the target. Each case has paired MCP-style and natural-language action
descriptions, sharing its context, reasoning, action order, and target.
Table~\ref{tab:alignment-dataset} reports the dataset sizes.

For each example, we first fix the permissible actions $\mathcal A_t$
and target action $a\in\mathcal A_t$. Each set contains 4--10 actions from
one domain. We balance set sizes and target frequencies within each domain,
approximately balance the use of other candidates, and shuffle their order.
GPT-5.5~\citep{openai2026gpt55} then generates task context and reasoning
conditioned on these actions and the target. The action descriptions provide
$\mathrm{desc}(e)$, while the generated context and reasoning enter $h_t$.
These examples supervise action selection without environment rollouts.

\paragraph{Inputs and Supervision.}
The LLM input combines the action descriptions, task context, and reasoning
with format instructions, an example call, and a prefix marking the
action-selection position. The example call uses an action sampled uniformly
from $\mathcal A_t$, without conditioning on the target.
Each input to $g$ contains the same action descriptions, the definition of
the action being encoded, and an instruction identifying that action.
It excludes the task context and reasoning and does not identify the
supervised target. All candidates are encoded as $\mathbf z_a$ and scored
against $\mathbf q_t$ at the action-selection position. We minimize
Equation~\ref{eq:dyad-alignment}, averaged over training examples, with a
softmax temperature of one. Only the projector is updated, and no
next-token prediction loss is applied to the supplied text.

\paragraph{Input Formats and Data Splits.}
Table~\ref{tab:alignment-dataset} summarizes the dataset sizes.
Each case has two versions, using structured MCP-style tool definitions or
natural-language action descriptions. They share the task context,
reasoning, candidate order, and target action, and remain in the same split.

\begin{table}[t]
    \centering
    \caption{\textbf{Alignment dataset.} Each case has paired MCP-style and natural-language records.}
    \label{tab:alignment-dataset}

    \setlength{\tabcolsep}{7pt}
    \renewcommand{\arraystretch}{1.12}
    \begin{tabular}{@{}lcccc@{}}
        \toprule
        & & \multicolumn{3}{c}{\textbf{Records}} \\
        \cmidrule(lr){3-5}
        \textbf{Split} & \textbf{Cases} & \textbf{MCP-style} & \textbf{Natural language} & \textbf{Total} \\
        \midrule
        Train & 1,050 & 1,050 & 1,050 & 2,100 \\
        Validation & 140 & 140 & 140 & 280 \\
        Test & 140 & 140 & 140 & 280 \\
        \midrule
        Total & 1,330 & 1,330 & 1,330 & 2,660 \\
        \bottomrule
    \end{tabular}
\end{table}

Validation and test are split by case, stratified by domain, target action,
and number of permissible actions, using seed 42. Their targets come from
the same 20 actions excluded from all training candidate sets; other
candidates may include both training and held-out actions. Thus, validation
and test contain distinct cases within a shared action pool.
We select the projector checkpoint by validation negative log-likelihood.
Because this evaluation pool had been used for model selection before
subdivision, the test split provides a diagnostic rather than a historically
independent evaluation.
\paragraph{Supervision and Data Splits.}
Only the projector is trained using Equation~\ref{eq:dyad-alignment};
the generated text is supplied as input, with supervision applied to the
target action. Validation and test targets come from actions absent from
all training candidate sets. The two splits contain distinct cases from
the same held-out action pool, with paired formats kept together.
We select checkpoints by validation cross-entropy. The evaluation pool
was used in earlier model selection, so its test split is not a
historically independent holdout.

\paragraph{Data Checks.}
We check action-definition structure, candidate coverage, paired-format
consistency, and split isolation. Near-duplicate contexts are identified by
word-set overlap within each domain and generation pool and regenerated.
These checks do not establish semantic equivalence of every description
pair or disjointness from downstream benchmark actions.


\subsection{Agentic Reinforcement Learning}
\label{app:agentic_rl}

\paragraph{Sampled Decisions and Loss Aggregation.}
Policy-gradient terms are computed for sampled vocabulary tokens and typed
actions. The fixed text appended after an action selection records that
decision and contributes no additional policy-gradient terms. Environment
observations are also excluded.
Let $i$ index a sampled decision and $p_i(\theta)$ denote its probability
under $\pi_\theta^{\mathrm{LM}}$ or $\pi_\theta^{\mathrm{act}}$, evaluated at
the recorded history and, for a typed action, its permissible action set.
The rollout parameters $\theta_{\mathrm{old}}$ and the advantage
$\widehat A_i$ computed from environment rewards are held fixed during each
update. The standard clipped surrogate for either decision type is
\begin{equation}
\label{eq:dyad-decision-loss}
\begin{aligned}
\rho_i(\theta)&=\frac{p_i(\theta)}{p_i(\theta_{\mathrm{old}})},\\
\ell_i(\theta)&=-\min\!\left(
\rho_i(\theta)\widehat A_i,\,
\operatorname{clip}\!\left(\rho_i(\theta),
1-\epsilon_{\mathrm{low}},1+\epsilon_{\mathrm{high}}\right)\widehat A_i
\right).
\end{aligned}
\end{equation}
The clipping bounds follow Table~\ref{tab:hyperparameters}.
Rollout and training probabilities use the same sampling temperature and
permissible actions. GRPO and GiGPO differ in how environment rewards are
used to estimate advantages~\citep{shao2024deepseekmath,feng2025gigpo}.

For aggregation group $b$, let $\mathcal I_b^{\mathrm{LM}}$ and
$\mathcal I_b^{\mathrm{act}}$ index sampled vocabulary tokens and typed
actions, respectively. Let $N_b$ be the number of response tokens in that
group and $c_b$ the weight assigned to that group. We aggregate the
sampled-decision losses into language and action contributions: 
\begin{equation}
\label{eq:dyad-loss-aggregation}
\begin{aligned}
\mathcal L^{\mathrm{LM}}
&=\sum_b\frac{c_b}{N_b}
\sum_{i\in\mathcal I_b^{\mathrm{LM}}}\ell_i(\theta),\\
\mathcal L^{\mathrm{act}}
&=\sum_b\frac{c_b}{N_b}
\sum_{i\in\mathcal I_b^{\mathrm{act}}}\ell_i(\theta).
\end{aligned}
\end{equation}
Both contributions share the same denominator and group weights.
The response-token count $N_b$ includes fixed action text but excludes
padding and environment observations. The action contribution is not
separately normalized by the number of actions.


In joint optimization, the policy-gradient update to $f$ and
$\mathbf W_{\mathrm{LM}}$ uses
$\mathcal L^{\mathrm{LM}}+\mathcal L^{\mathrm{act}}$, with action
representations $\mathbf z_a$ treated as constants. The update to $g$
uses $\mathcal L^{\mathrm{act}}$, with state representations
$\mathbf q_t$ treated as constants, and trains both its backbone and
projector. The action contribution has the same numerical value in both
updates, but its gradients are taken with respect to different components.
In frozen-LLM adaptation, only the update to $g$ is applied.

\section{Experimental Details}
\label{app:experimental_details}


\subsection{Agentic Environments}
\label{app:agentic_environments}

Table~\ref{tab:environment_statistics} summarizes task counts and the average
number of available tool or operation types per task. For ALFWorld and
WebShop, these counts exclude state-dependent arguments and therefore do
not measure the number of permissible choices at each decision step.
The substantially larger toolsets in DIVE contrast with CodeGym's smaller
sets of atomic functions. SWE exposes only one Bash interface, but its
command arguments support repository inspection, editing, and testing.
These statistics characterize the breadth of exposed interfaces;
the number of interfaces alone does not determine the complexity of
choosing and executing an action.

\begin{table}[t]
    \centering
    \caption{\textbf{Environment statistics.} Environments cover training and transfer evaluation.}
    \label{tab:environment_statistics}
    \setlength{\tabcolsep}{8pt}
    \renewcommand{\arraystretch}{1.12}
    \begin{tabular}{@{}lcc@{}}
        \toprule
        \textbf{Environment} & \textbf{\# Tasks} & \textbf{Avg. tools / actions} \\
        \midrule
        DIVE & 1,499 & 40.87 \\
        CodeGym & 80,130 & 6.48 \\
        ALFWorld & 3,827 & 14 \\
        WebShop & 11,687 & 2 \\
        $\tau^2$-bench & 278 & 15.41 \\
        SWE & 500 & 1 \\
        \bottomrule
    \end{tabular}
\end{table}

\paragraph{DIVE~\citep{chen2026dive}.}
The dataset contains diverse tool-use tasks synthesized from execution traces of real-world tools. Each task pairs a query with an available toolset and a verifiable reference answer, requiring agents to gather and process information through multiple tool calls.

\paragraph{CodeGym~\citep{du2025codegym}.}
Coding problems are converted into interactive tool-use environments by exposing atomic functions from their solutions as callable tools. Agents solve tasks through sequential tool interactions, while the underlying code and test cases support automatic verification.
The counts in Table~\ref{tab:environment_statistics} use the English-task
reproduction released as \href{https://huggingface.co/datasets/VanishD/CodeGym/tree/85286359a342f7a288aea74273772b69b9b784c2}{VanishD/CodeGym}.

\paragraph{ALFWorld~\citep{shridhar2020alfworld}.}
This benchmark provides text-based environments for household tasks, such as finding, placing, cleaning, heating, and cooling objects. Agents navigate and manipulate objects through textual commands, using observations from the environment to track progress toward the task goal.

\paragraph{WebShop~\citep{yao2022webshop}.}
In this simulated e-commerce environment, agents follow natural-language shopping instructions. They search for products, inspect product details, select options, and make a purchase that satisfies the requested attributes and constraints.

\paragraph{$\tau^2$-bench~\citep{barres2025tau2bench}.}
This benchmark evaluates conversational agents that use tools and communicate with users to complete tasks. Its dual-control telecom environment allows both the agent and the user to act on a shared environment state, testing coordination alongside tool use and reasoning.
The statistics use the Retail, Airline, and Telecom base splits at
\href{https://github.com/sierra-research/tau2-bench/tree/a2c024725189473d2d7cea3a5cfdbcc67478e41f}{repository revision \texttt{a2c0247}},
which includes subsequent $\tau^3$ updates.

\paragraph{SWE-bench Verified.}
This human-filtered subset of SWE-bench~\citep{jimenez2023swebench} evaluates the resolution of real-world GitHub issues.\footnote{\url{https://www.swebench.com/}} Given an issue description and a code repository, agents inspect and modify the code to produce a patch, whose correctness is assessed using the benchmark's tests.
We evaluate a 100-task subset of the 500 verified instances.
We use the Bash interface of mini-swe-agent.\footnote{\url{https://github.com/SWE-agent/mini-swe-agent}}
The evaluated model directly generates shell commands for repository inspection,
editing, and testing. A GiGPO-style prompt provides recent observation--action
pairs and the current execution feedback at each step.
The same Bash interface submits the patch and terminates the episode.

The cross-environment evaluations use agents trained with RL on DIVE or
CodeGym without further RL on ALFWorld, WebShop, $\tau^2$-bench, or
SWE-bench Verified. This separation refers to the RL training environments;
the alignment-data split is described in Appendix~\ref{app:action_alignment}.
\subsection{General Capability Benchmarks}
\label{app:general_capability_benchmarks}

\paragraph{MMLU-Pro~\citep{wang2024mmlupro}.}
The benchmark extends MMLU with more challenging, reasoning-focused multiple-choice questions across academic subjects and up to ten answer choices per question. We evaluate on the full test set of 12,032 questions to assess general knowledge and reasoning after agentic post-training.

\paragraph{HMMT February 2026.}
We use the \href{https://huggingface.co/datasets/MathArena/hmmt_feb_2026}{MathArena release}~\citep{dekoninck2026matharena} of problems from the February 2026 Harvard--MIT Mathematics Tournament. These competition problems assess mathematical reasoning across algebra, geometry, number theory, and combinatorics.
We evaluate on all 33 problems and report avg@4, averaging correctness over
four generated answers per problem and then across problems.

\paragraph{LiveCodeBench v6~\citep{jain2024livecodebench}.}
The benchmark evaluates coding capabilities using programming contest problems collected over time from LeetCode, AtCoder, and Codeforces. We evaluate on the full version~6 test set of 1,055 problems, released from May 2023 to April 2025, with generated programs evaluated against the provided test cases.

\subsection{JevBench}
\label{app:jevbench}

\paragraph{Tasks and Evaluation Subset.}
JevBench evaluates typed decision-making given a state, instructions,
decision criteria, and a finite set of candidate answers.\footnote{\url{https://github.com/fstandhartinger/jevbench}}
Its tasks include categorical choices, yes/no judgments, and ordinal
ratings, covering problems such as routing, policy compliance,
information extraction, and numerical reasoning.
Each item evaluates a standalone decision without an environment rollout.
We use the public subset of 231 items: 48 Easy, 72 Standard, and 111 Hard.

\paragraph{Metrics and Latency Source.}
Table~\ref{tab:jevbench-results} reports accuracy within each difficulty
tier; Overall is the average weighted by the number of items per tier.
Mean output tokens include reasoning and final answers, and mean latency
measures time until the complete response is returned.
The Jev latency entry is recomputed from the benchmark's
\href{https://github.com/fstandhartinger/jevbench/blob/1bcc55eb6c8cffde2306b3db03ede39b61c6152a/results/v1.2/jevbench-v1.2-per-task.json}{published per-item API timings}
for the same public subset. These timings come from a separate evaluation
and include network overhead. They provide an API latency reference;
the different deployments do not establish a matched-hardware speed
comparison. A corresponding Jev output-token measurement for this subset
is unavailable.

\subsection{Hyperparameters}
\label{app:hyperparameters}

Table~\ref{tab:hyperparameters} summarizes the hyperparameters for agentic RL in the joint optimization setting.
The response limit bounds each turn, while the interaction limit bounds
the trajectory. Shorter individual responses can therefore coexist with
longer interaction sequences, as observed on CodeGym in
Appendix~\ref{app:additional_training_dynamics}. Since these budgets and
rollout batch sizes differ across environments, we interpret training
curves through comparisons within each environment and model scale.

\begin{table}[t]
    \centering
    \caption{\textbf{Agentic RL hyperparameters.} Settings correspond to joint post-training.}
    \label{tab:hyperparameters}
    \setlength{\tabcolsep}{3pt}
    \renewcommand{\arraystretch}{1.08}
    \begin{tabular*}{\linewidth}{@{\extracolsep{\fill}}lcccc@{}}
        \toprule
        \textbf{Environment} & \textbf{ALFWorld} & \textbf{WebShop} & \textbf{DIVE} & \textbf{CodeGym} \\
        \midrule
        \multicolumn{5}{c}{\textit{Interaction budgets}} \\
        \midrule
        Max prompt length (tokens) & 2,048 & 4,096 & 32,768 & 5,120 \\
        Max interaction steps & 50 & 15 & 30 & 80 \\
        Memory context window (steps) & 2 & 2 & 2 & 2 \\
        Max response length (tokens/turn) & 512 & 512 & 512 & 386 \\
        \midrule
        \multicolumn{5}{c}{\textit{Sampling and batching}} \\
        \midrule
        Task groups per rollout & 16 & 32 & 32 & 512 \\
        Trajectories per task group & 8 & 8 & 8 & 8 \\
        Update mini-batch size (steps) & 256 & 64 & 32 & 64 \\
        Training budget & 2 epochs & 2 epochs & 2 epochs & 3 epochs \\
        Temperature (training) & 1.0 & 1.0 & 0.9 & 1.0 \\
        Temperature (evaluation) & 0.4 & 0.4 & 0.4 & 0.4 \\
        Top-$p$ (evaluation) & 1.0 & 1.0 & 1.0 & 1.0 \\
        \midrule
        \multicolumn{5}{c}{\textit{Optimization}} \\
        \midrule
        Policy learning rate & $10^{-6}$ & $10^{-6}$ & $10^{-6}$ & $10^{-6}$ \\
        PPO clip range $(\epsilon_{\mathrm{low}},\epsilon_{\mathrm{high}})$ & $(0.2,0.2)$ & $(0.2,0.2)$ & $(0.2,0.2)$ & $(0.2,0.2)$ \\
        KL loss coefficient & 0.01 & 0.01 & 0.001 & 0 \\
        \midrule
        \multicolumn{5}{c}{\textit{Method-specific settings}} \\
        \midrule
        GiGPO discount factor $\gamma$ & 0.95 & 0.95 & 0.95 & 0.95 \\
        GiGPO step advantage weight $\omega$ & 1.0 & 1.0 & 1.0 & 1.0 \\
        \bottomrule
    \end{tabular*}
    \par\smallskip
    \begin{minipage}{\linewidth}
        \footnotesize
        \textit{Notes.}
        Prompt and response limits apply per interaction; responses exclude tool outputs.
        Mini-batches count interaction steps. Thinking mode is disabled.
    \end{minipage}
\end{table}

\begingroup
\raggedbottom
\subsection{Prompt Templates}
\label{app:prompt_templates}

Figures~\ref{fig:prompt_dive}--\ref{fig:prompt_swebench} summarize the
environment prompt templates. Named placeholders represent the task,
available actions, observations, and interaction history. The templates
request explicit reasoning; disabling the model's native thinking mode
does not remove these instructions.

For templates with explicit history fields, these contain recent observation--action pairs, while the current observation supplies the latest feedback. When no history is included, the corresponding history text is omitted; DIVE, ALFWorld, and WebShop also omit step-count text. The response instructions remain unchanged. WebShop additionally falls back to its no-history template when a prompt with history exceeds 13,000 characters.
SWE-bench Verified follows the same GiGPO-style per-step structure, with
fixed repository and submission instructions in the system message.
The user message contains the task, recent history, current observation,
and available tool. We retain our \texttt{<think>} and \texttt{<action>}
response format while using mini-swe-agent's Bash interface.

\begin{figure}[t]
\centering
\begin{PromptTemplate}{DIVE prompt template}
\PromptRole{User}
You are an expert agent solving a task using the available tools.\par
Your task is to: \PromptVar{task_description}\par
Prior to this step, you have already taken \PromptVar{step_count} step(s). Below are the most recent \PromptVar{history_length} observations and the corresponding actions you took: \PromptVar{action_history}\par
You are now at step \PromptVar{current_step} and your current observation is: \PromptVar{current_observation}\par
Your available tools are: \PromptVar{available_actions}\par
Use these tool schemas to construct calls; the interface specifies the native tool-call syntax inside \PromptToken{<action>}.\par
\PromptGap
Reason about the next step within \PromptToken{<think>} \PromptToken{</think>} tags.\par
Then enclose your action within \PromptToken{<action>} \PromptToken{</action>} tags.\par
Inside \PromptToken{<action>}, call the appropriate tools using the native tool-call format and their declared arguments, then close \PromptToken{</action>} and wait for their results.\par
When ready, put your final answer inside \PromptToken{<action>} \PromptToken{</action>} instead of tool calls, in the format requested by the query.\par
An action without tool calls ends the task; do not output reasoning alone or text outside these two blocks.\par
\end{PromptTemplate}
\caption{\textbf{Prompt template for DIVE.} It specifies how the agent calls tools and submits its final answer.}
\label{fig:prompt_dive}
\end{figure}

\begin{figure}[t]
\centering
\begin{PromptTemplate}{CodeGym prompt template}
\PromptRole{System}
\PromptVar{original_system_messages}\par
\PromptRole{User}
You are an expert agent operating in an interactive function-calling environment.\par
Your task is to: \PromptVar{task_description}\par
Prior to this step, you have already taken \PromptVar{step_count} step(s).\par
Below are the most recent \PromptVar{history_length} observations and the corresponding actions you took: \PromptVar{action_history}\par
You are now at step \PromptVar{current_step} and your current observation is: \PromptVar{current_observation}\par
Your available functions are: [\PromptVar{available_actions}].\par
\PromptGap
Reason about the next step within \PromptToken{<think>} \PromptToken{</think>} tags.\par
Then output exactly one function call using its declared arguments:\par
{\ttfamily \PromptToken{<|FunctionCallBegin|>}\allowbreak[\{"name": "<function name>", "parameters": \{"<argument name>": "<value>"\}\}]\allowbreak\PromptToken{<|FunctionCallEnd|>}\par}
Use \{\} for no arguments; do not write executable code.\par
Wait for feedback after each call and follow the task's submission rules.\par
\end{PromptTemplate}
\caption{\textbf{Prompt template for CodeGym.} The agent issues one function call at a time and waits for feedback before continuing.}
\label{fig:prompt_codegym}
\end{figure}

\begin{figure}[t]
\centering
\begin{PromptTemplate}{ALFWorld prompt template}
\PromptRole{User}
You are an expert agent operating in an interactive household environment. Your task is to: \PromptVar{task_description}\par
Prior to this step, you have already taken \PromptVar{step_count} step(s). Below are the most recent \PromptVar{history_length} observations and the corresponding actions you took: \PromptVar{action_history}\par
You are now at step \PromptVar{current_step} and your current observation is: \PromptVar{current_observation}\par
Your admissible actions of the current situation are: [\PromptVar{admissible_actions}].\par
\PromptGap
Now it's your turn to take an action.\par
You should first reason step-by-step about the current situation. This reasoning process MUST be enclosed within \PromptToken{<think>} \PromptToken{</think>} tags.\par
Once you've finished your reasoning, you should choose an admissible action for current step and present it within \PromptToken{<action>} \PromptToken{</action>} tags.\par
\end{PromptTemplate}
\caption{\textbf{Prompt template for ALFWorld.} The agent selects an admissible action based on the current observation and interaction history.}
\label{fig:prompt_alfworld}
\end{figure}

\begin{figure}[t]
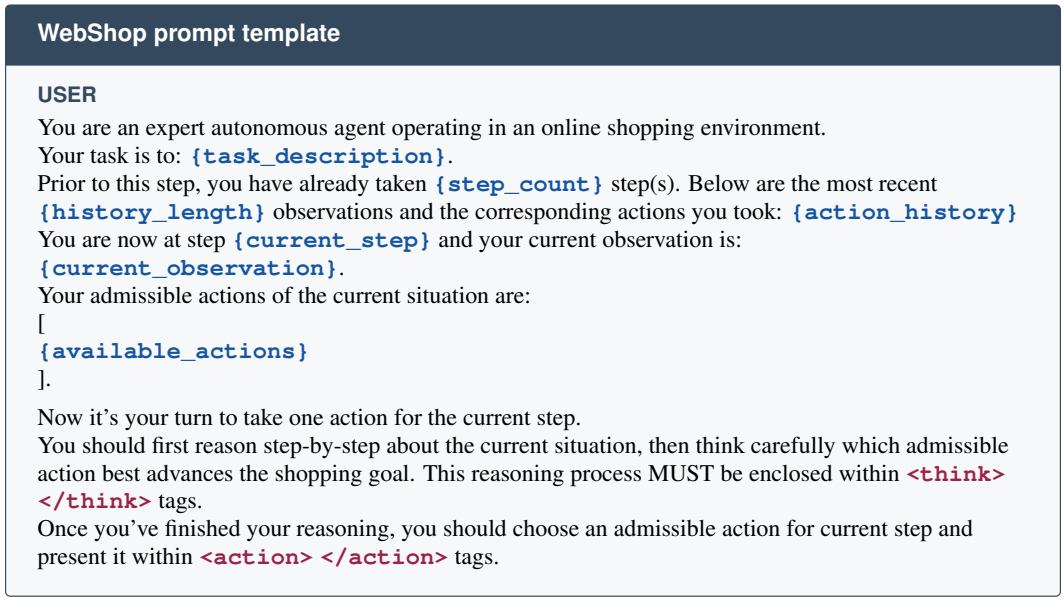

\centering
\begin{PromptTemplate}{WebShop prompt template}
\PromptRole{User}
You are an expert autonomous agent operating in an online shopping environment.\par
Your task is to: \PromptVar{task_description}.\par
Prior to this step, you have already taken \PromptVar{step_count} step(s). Below are the most recent \PromptVar{history_length} observations and the corresponding actions you took: \PromptVar{action_history}\par
You are now at step \PromptVar{current_step} and your current observation is: \PromptVar{current_observation}.\par
Your admissible actions of the current situation are:\par
[\par
\PromptVar{available_actions}\par
].\par
\PromptGap
Now it's your turn to take one action for the current step.\par
You should first reason step-by-step about the current situation, then think carefully which admissible action best advances the shopping goal. This reasoning process MUST be enclosed within \PromptToken{<think>} \PromptToken{</think>} tags.\par
Once you've finished your reasoning, you should choose an admissible action for current step and present it within \PromptToken{<action>} \PromptToken{</action>} tags.\par
\end{PromptTemplate}
\caption{\textbf{Prompt template for WebShop.} The agent uses the shopping instruction and current observation to select its next action.}
\label{fig:prompt_webshop}
\end{figure}

\begin{figure}[t]
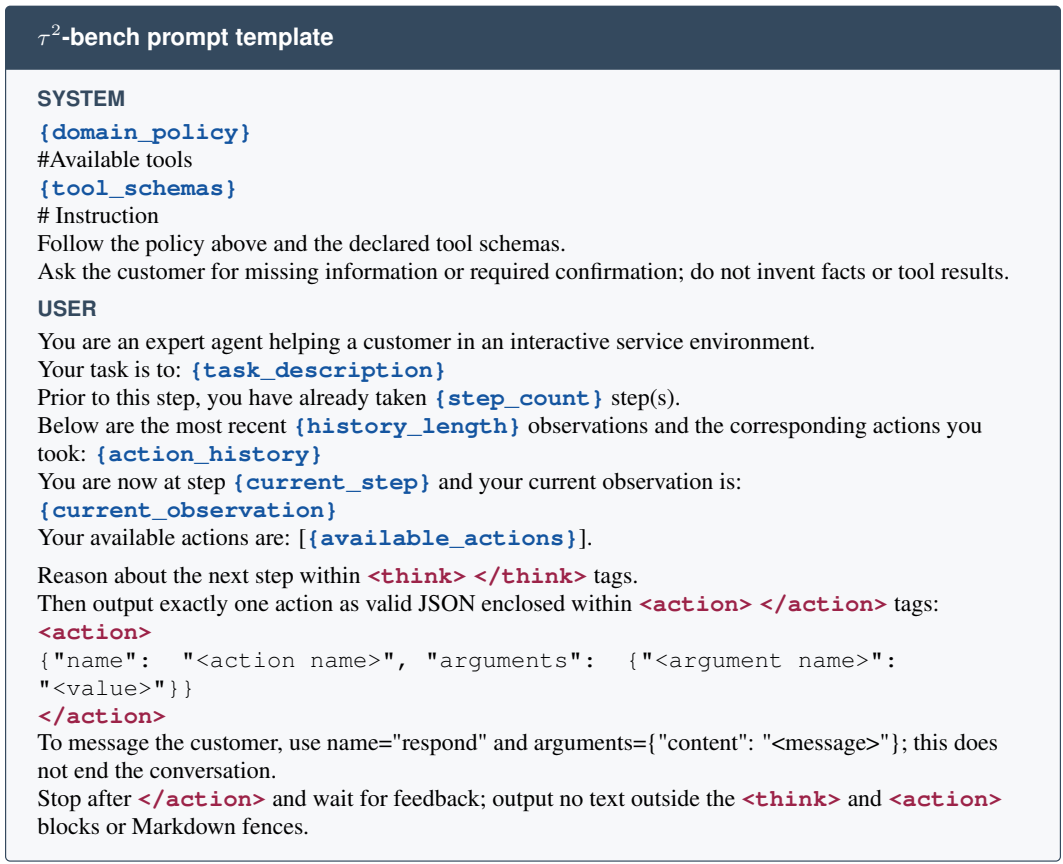

\centering
\begin{PromptTemplate}{$\tau^2$-bench prompt template}
\PromptRole{System}
\PromptVar{domain_policy}\par
\#Available tools\par
\PromptVar{tool_schemas}\par
\# Instruction\par
Follow the policy above and the declared tool schemas.\par
Ask the customer for missing information or required confirmation; do not invent facts or tool results.\par
\PromptRole{User}
You are an expert agent helping a customer in an interactive service environment.\par
Your task is to: \PromptVar{task_description}\par
Prior to this step, you have already taken \PromptVar{step_count} step(s).\par
Below are the most recent \PromptVar{history_length} observations and the corresponding actions you took: \PromptVar{action_history}\par
You are now at step \PromptVar{current_step} and your current observation is: \PromptVar{current_observation}\par
Your available actions are: [\PromptVar{available_actions}].\par
\PromptGap
Reason about the next step within \PromptToken{<think>} \PromptToken{</think>} tags.\par
Then output exactly one action as valid JSON enclosed within \PromptToken{<action>} \PromptToken{</action>} tags:\par
\PromptToken{<action>}\par
{\ttfamily \{"name": "<action name>", "arguments": \{"<argument name>": "<value>"\}\}\par}
\PromptToken{</action>}\par
To message the customer, use name="respond" and arguments=\{"content": "<message>"\}; this does not end the conversation.\par
Stop after \PromptToken{</action>} and wait for feedback; output no text outside the \PromptToken{<think>} and \PromptToken{<action>} blocks or Markdown fences.\par
\end{PromptTemplate}
\caption{\textbf{Prompt template for $\tau^2$-bench.} The agent follows the domain policy when using tools and communicating with the customer.}
\label{fig:prompt_t2bench}
\end{figure}

\begin{figure}[t]
\centering
\begin{PromptTemplate}{SWE-bench Verified prompt template}
\PromptRole{System}
Fix the issue in /testbed and verify the fix. Each command runs in a fresh shell. Do not modify tests or configuration files, or commit changes.\par
\PromptGap
Save a source-only \texttt{git diff} to \texttt{patch.txt}, using \texttt{git add -N} for new files. Exclude reproduction scripts, helpers, and binaries. Inspect the patch in a separate call, then submit in a final bash call:\par
\texttt{echo COMPLETE\_TASK\_AND\_SUBMIT\_FINAL\_OUTPUT \&\& cat patch.txt}\par
Exit status 0 completes submission.\par
\PromptRole{User}
Task: \PromptVar{task_description}\par
Recent observation--action history: \PromptVar{action_history}\par
Current step: \PromptVar{current_step}\par
Current observation: \PromptVar{current_observation}\par
Available tools: [\PromptVar{available_actions}].\par
\PromptGap
Put reasoning inside \PromptToken{<think>} \PromptToken{</think>}, followed by one \PromptToken{<action>} \PromptToken{</action>} block with native bash calls. Include at least one call; group only independent commands. Output nothing outside these blocks; then stop and wait for results.\par
\end{PromptTemplate}
\caption{\textbf{Prompt template for SWE-bench Verified.} GiGPO-style context with mini-swe-agent's Bash interface.}
\label{fig:prompt_swebench}
\end{figure}

\par\endgroup
\FloatBarrier

\section{Additional Results and Analyses}
\label{app:additional_results}

\subsection{Task-Specific Post-training on ALFWorld and WebShop}
\label{app:task_specific_posttraining}

We train separate agents on ALFWorld and WebShop, jointly updating the LLM
and action encoder for Dyad. Table~\ref{tab:task-specific-capability} reports
general capabilities for the same agents evaluated in
Table~\ref{tab:task-specific-performance}. These runs are separate from the
DIVE- and CodeGym-trained agents used for cross-environment evaluation.

\begin{table}[t]
    \centering
    \caption{\textbf{General capability retention.} Models are post-trained on ALFWorld or WebShop.}
    \label{tab:task-specific-capability}
    \setlength{\tabcolsep}{2.0pt}
    \renewcommand{\arraystretch}{1.0}
    \begin{tabular*}{\linewidth}{@{\extracolsep{\fill}}l*{3}{c}@{\hspace{2pt}}*{3}{c}@{}}
        \toprule
         & \multicolumn{3}{c}{\textbf{ALFWorld}} & \multicolumn{3}{c}{\textbf{WebShop}} \\
        \cmidrule(lr){2-4}\cmidrule(l){5-7}
        Model / Method & MMLU-Pro & HMMT & LiveCodeBench & MMLU-Pro & HMMT & LiveCodeBench \\
        \midrule
        \textit{Qwen3.5-4B} & \textbf{69.34} & 36.26 & 47.39 & 69.34 & \textbf{36.26} & 47.39 \\
        \addlinespace[1pt]
        \quad + GRPO & 68.08 & 30.33 & 46.21 & 67.93 & 33.30 & 46.62 \\
        \quad + Dyad-GRPO & 68.96 & 33.36 & 47.02 & 69.02 & 34.76 & 47.11 \\
        \addlinespace[1pt]
        \quad + GiGPO & 67.62 & 32.46 & 46.78 & 67.51 & 33.05 & 46.06 \\
        \quad + Dyad-GiGPO & 68.10 & \textbf{36.33} & \textbf{47.52} & \textbf{69.53} & 35.16 & \textbf{47.89} \\
        \midrule
        \textit{Qwen3.5-9B} & 71.88 & 39.75 & 59.62 & \textbf{71.88} & \textbf{39.75} & 59.62 \\
        \addlinespace[1pt]
        \quad + GRPO & 70.71 & 35.13 & 58.32 & 70.39 & 36.66 & 57.81 \\
        \quad + Dyad-GRPO & \textbf{71.92} & \textbf{42.90} & \textbf{60.89} & 71.84 & 36.98 & 59.67 \\
        \addlinespace[1pt]
        \quad + GiGPO & 70.02 & 37.06 & 58.06 & 70.89 & 37.34 & 58.42 \\
        \quad + Dyad-GiGPO & 71.21 & 40.29 & 59.99 & 71.42 & 38.09 & \textbf{60.10} \\
        \bottomrule
    \end{tabular*}
\end{table}

\noindent\textbf{Comparison with RL Baselines.}
Both Dyad variants outperform their corresponding RL baselines in all
24 comparisons across the two model scales, two training environments,
and three capability benchmarks. The advantage extends to general
knowledge, mathematical reasoning, and coding. Together with the
consistently higher in-domain
scores in Table~\ref{tab:task-specific-performance}, these results show
that stronger task adaptation with Dyad can accompany better general
capabilities than conventional RL post-training.

\noindent\textbf{Retention Relative to the Original Models.}
GRPO and GiGPO score below the original models in every reported
capability evaluation. Dyad reduces these losses, but recovery varies
with the training environment and capability. At 9B, both Dyad variants
exceed the original model's coding score after training on either
environment. Mathematical reasoning is more dependent on the training
source: both 9B Dyad variants exceed the original HMMT score after
ALFWorld training, whereas both remain below it after WebShop training.
At 4B, several scores also remain below their original levels despite
improving over the RL baselines. Thus, the paired advantage supports
better capability retention during task-specific joint training, although
it does not eliminate degradation in every setting.

\subsection{Additional Training Dynamics}
\label{app:additional_training_dynamics}

\begin{figure}[t]
    \centering
    {\small\textbf{Qwen3.5-4B}\par}\smallskip
    \includegraphics[width=\linewidth]{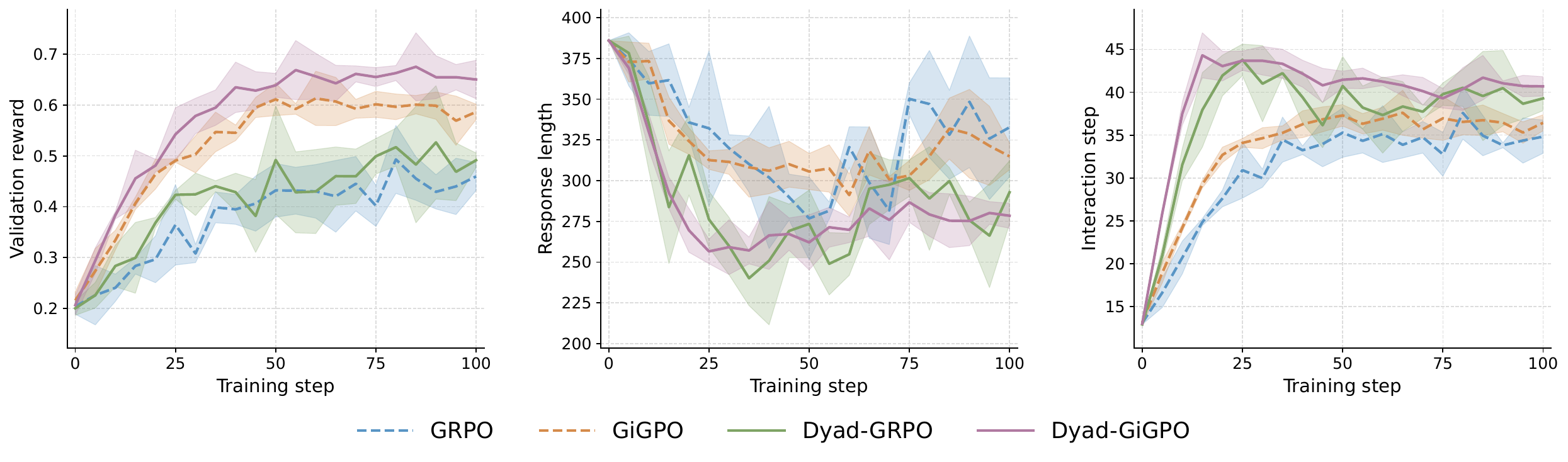}
    \par\medskip
    {\small\textbf{Qwen3.5-27B}\par}\smallskip
    \includegraphics[width=\linewidth]{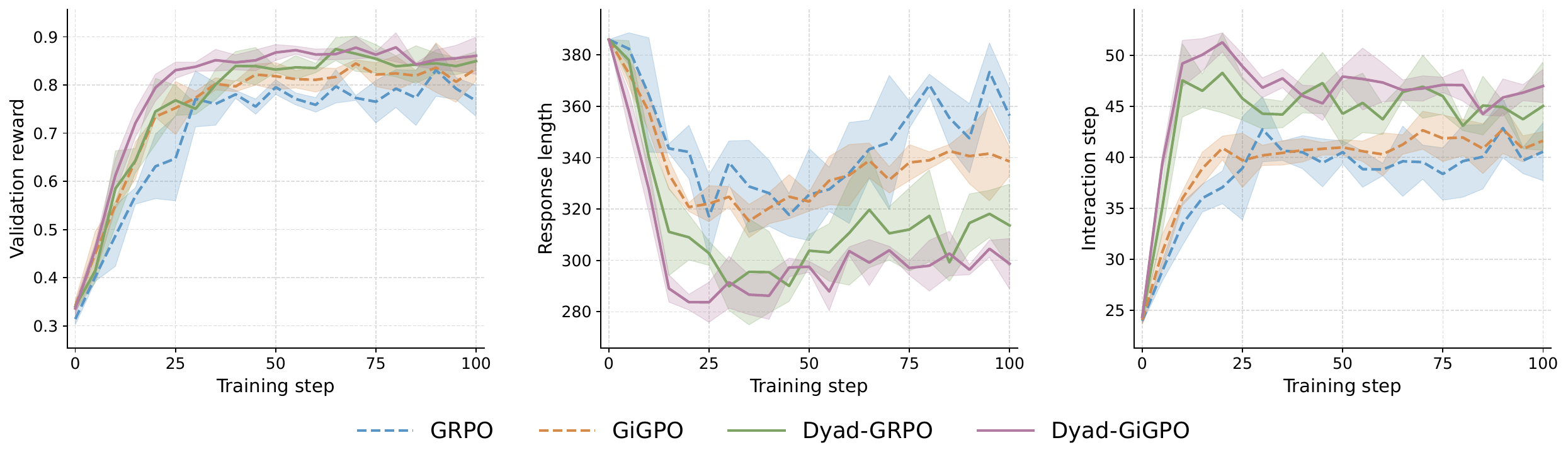}
    \caption{\textbf{Additional training dynamics on CodeGym.} Validation reward, response length, and environment interactions for Qwen3.5-4B (top) and Qwen3.5-27B (bottom).}
    \label{fig:codegym-other-scales-training}
\end{figure}

\noindent\textbf{CodeGym.}
Figure~\ref{fig:codegym-other-scales-training} extends the 9B results in
Section~\ref{sec:agentic_effectiveness} to Qwen3.5-4B and Qwen3.5-27B.
At both scales, Dyad-GRPO and Dyad-GiGPO achieve higher final validation
rewards than their corresponding baselines, with generally shorter
responses and more environment interactions. Despite greater fluctuations
at 4B, the paired advantage remains visible late in training.

\begin{figure}[t]
    \centering
    {\small\textbf{Qwen3.5-4B}\par}\smallskip
    \includegraphics[width=\linewidth]{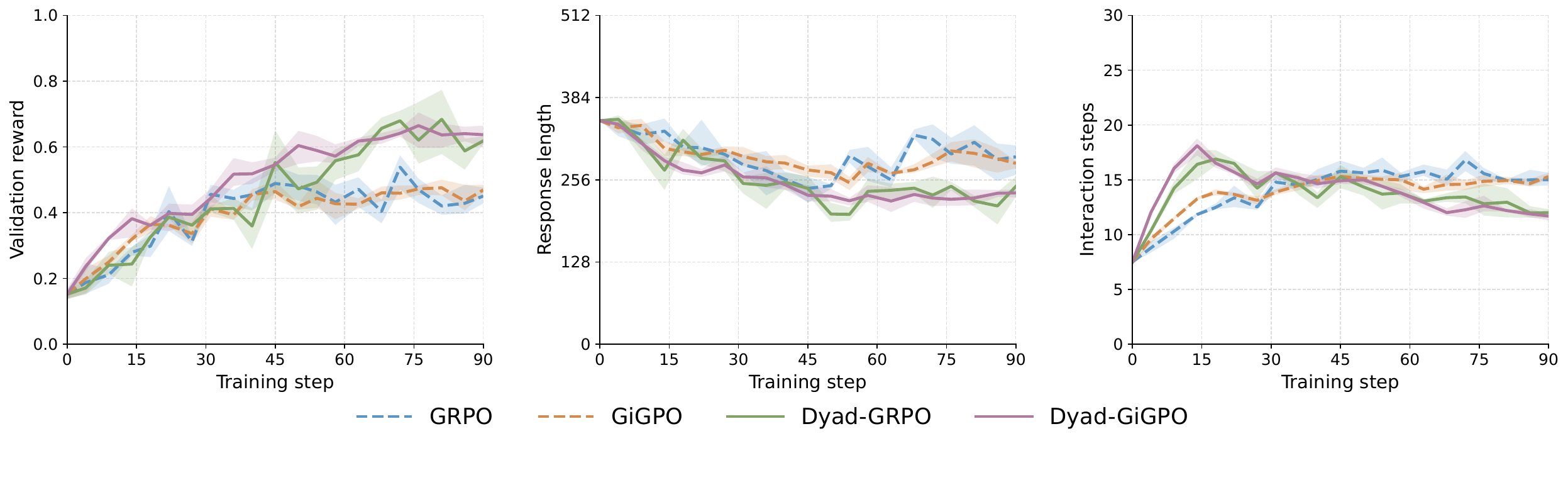}
    \par\medskip
    {\small\textbf{Qwen3.5-9B}\par}\smallskip
    \includegraphics[width=\linewidth]{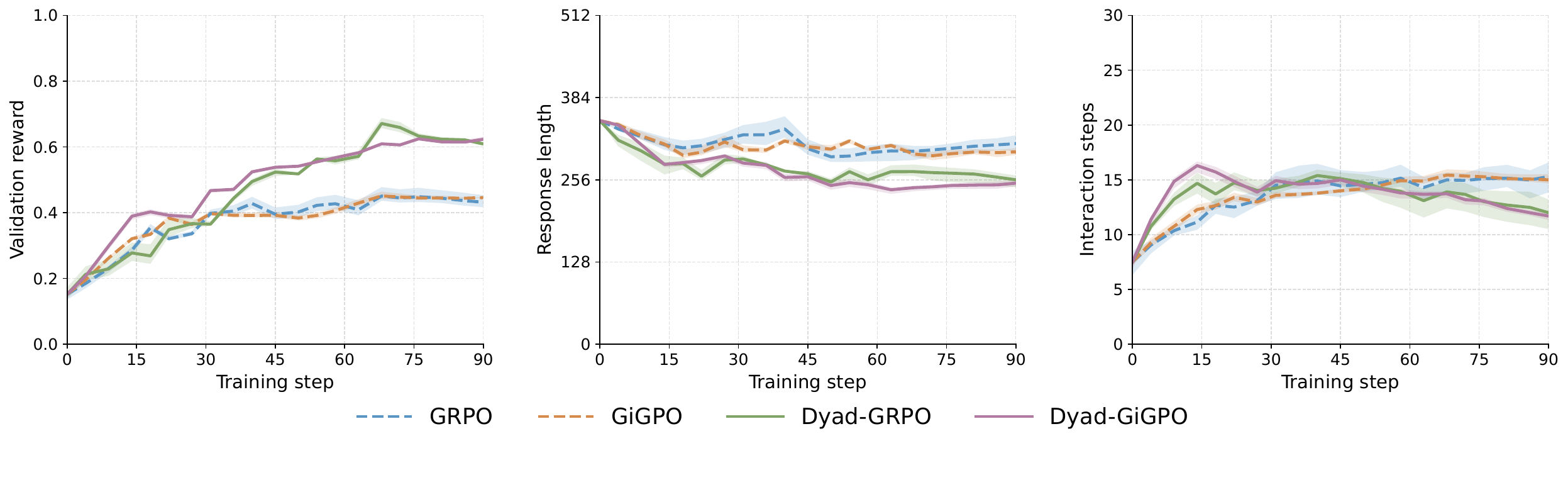}
    \par\medskip
    {\small\textbf{Qwen3.5-27B}\par}\smallskip
    \includegraphics[width=\linewidth]{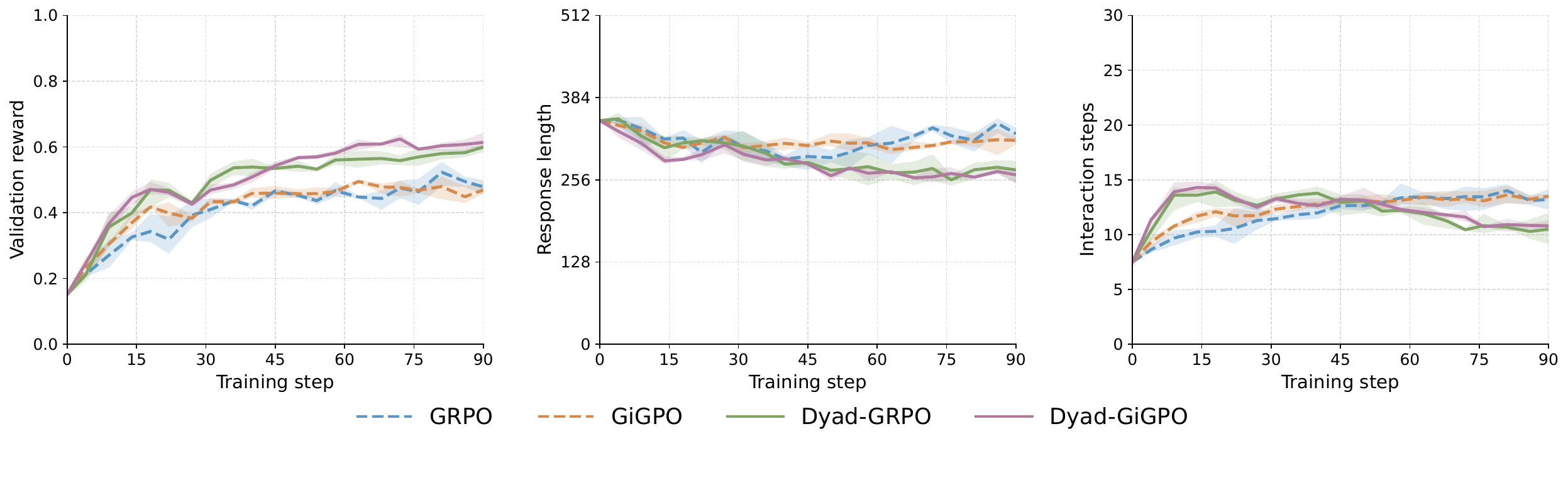}
    \caption{\textbf{Training dynamics on DIVE.} Models are jointly post-trained on DIVE.}
    \label{fig:dive-training}
\end{figure}

\noindent\textbf{DIVE.}
Figure~\ref{fig:dive-training} reports training dynamics for
Qwen3.5-4B, Qwen3.5-9B, and Qwen3.5-27B. Both Dyad variants attain higher
final validation rewards and shorter responses than their corresponding
baselines. Their interaction counts rise initially but decline later,
finishing below the baselines. The reward advantage is therefore consistent
across the two environments, while the interaction pattern differs:
more interactions on CodeGym and fewer on DIVE.

\FloatBarrier

\end{document}